\documentclass{article}

\usepackage{hyperref}

\usepackage{subcaption}
\usepackage{amsfonts}
\usepackage{float}
\usepackage[a4paper,tmargin=30mm,bmargin=30mm,lmargin=15mm,rmargin=15mm,headsep=20mm]{geometry}

\usepackage[acronym, shortcuts]{glossaries}
\usepackage{tikz}
\makeglossaries

\newacronym{ai}{AI}{Artificial Intelligence}
\newacronym{ann}{ANN}{Artificial Neural Network}
\newacronym{dl}{DL}{Deep Learning}
\newacronym{ae}{AE}{Autoencoder}
\newacronym{vae}{VAE}{Variational Autoencoder}
\newacronym{ml}{ML}{Machine Learning}
\newacronym{nn}{NN}{Neural Network}
\newacronym{g}{G}{Generator}
\newacronym{d}{D}{Discriminator}
\newacronym{da}{DA}{Data Augmentation}
\newacronym{nlp}{NLP}{Natural Language Processing}
\newacronym{mae}{MAE}{Mean Absolute Error}
\newacronym{tsne}{t-SNE}{$t$-distributed Stochastic Neighbour Embedding}
\newacronym{pca}{PCA}{Principal Component Analysis}
\newacronym{kldivergence}{KL Divergence}{Kullback-Leibler Divergence}
\newacronym{mpce}{MPCE}{Mean Per-Class Error}
\newacronym{lstm}{LSTM}{Long-Short Term Memory}
\newacronym{spawner}{SPAWNER}{SuboPtimAl Warped time series geNEratoR}
\newacronym{dtw}{DTW}{Dynamic Time Warping}
\newacronym{cnn}{CNN}{Convolutional Neural Network}
\newacronym{svm}{SVM}{Support Vector Machine}
\newacronym{ecg}{ECG}{Electrocardiogram}
\newacronym{eeg}{EEG}{Electroencephalogram}
\newacronym{gasf}{GASF}{Gramian Angular Summation Field}
\newacronym{cvae}{CVAE}{Conditional \gls{vae}}
\newacronym{sisvae}{SISVAE}{Smoothness-Inducing Sequential \gls{vae}}
\newacronym{JSD}{JSD}{Jensen-Shannon Divergence}
\newacronym{tstr}{TSTR}{Train Synthetic, Test Real}

\newacronym{gan}{GAN}{Generative Adversarial Network}
\newacronym{timegan}{TimeGAN}{time series \gls{gan}}
\newacronym{acgan}{ACGAN}{Auxiliary Classifier \gls{gan}}
\newacronym{rcgan}{RCGAN}{Recurrent Conditional \gls{gan}}
\newacronym{tcgan}{T-CGAN}{Temporal-Conditional \gls{gan}}
\newacronym{cgan}{CGAN}{Conditional \gls{gan}}
\newacronym{dcgan}{DCGAN}{Deep Convolutional \gls{gan}}
\newacronym{tsdigan}{TSDIGAN}{Traffic Sensor Data Imputation \gls{gan}}
\newacronym{crnngan}{C-RNN-GAN}{Continuous recurrent \gls{gan}}
\newacronym{mffgan}{MFF-GAN}{MultiLevel Feature Fusion with \gls{gan}}
\newacronym{wgan}{WGAN}{Wasserstein \gls{gan}}

\begin{document}

\title{Data Augmentation techniques in time series domain: A survey and taxonomy}

\author{Guillermo Iglesias
\thanks{Departamento de Sistemas Inform\'aticos, Universidad Polit\'ecnica de Madrid, Spain}
\\{guillermo.iglesias.hernandez@alumnos.upm.es}
\and
Edgar Talavera
\footnotemark[1]
\\{e.talavera@upm.es}
\and
\'Angel Gonz\'alez-Prieto
\thanks{Departamento de \'Algebra, Geometr\'ia y Topolog\'ia, Universidad Complutense de Madrid, Spain}
\thanks{Instituto de Ciencias Matem\'aticas (CSIC-UAM-UCM-UC3M), Spain}
\\{angelgonzalezprieto@ucm.es}
\and
Alberto Mozo
\footnotemark[1]
\\{a.mozo@upm.es}
\and
Sandra G\'omez-Canaval
\footnotemark[1]
\\{sm.gomez@upm.es}}

\date{}

\maketitle


\abstract{With the latest advances in Deep Learning-based} generative models, it has not taken long to take advantage of their remarkable performance in the area of time series. Deep neural networks used to work with time series heavily depend on the size and consistency of the datasets used in training. These features are not usually abundant in the real world, where they are usually limited and often have constraints that must be guaranteed. Therefore, an effective way to increase the amount of data is by using Data Augmentation techniques, either by adding noise or permutations and by generating new synthetic data. This work systematically reviews the current state-of-the-art in the area to provide an overview of all available algorithms and proposes a taxonomy of the most relevant research. The efficiency of the different variants will be evaluated as a central part of the process, as well as the different metrics to evaluate the performance and the main problems concerning each model will be analysed. The ultimate aim of this study is to provide a summary of the evolution and performance of areas that produce better results to guide future researchers in this field.

\glsresetall

\section{Introduction}\label{sec1}

From the advent of \gls{dl}, a large part of research and the efforts of the scientific community have focused on solving and improving supervised training tasks. Supervised learning requires larger datasets, with a large number of different features, and these samples also need to be labelled in order to be feasible. With the latest models, it is becoming more difficult to obtain a suitable dataset, as the models require larger amounts of data to carry out the training. Usually, there exists a large number of public repositories from which to obtain suitable datasets for training in most application areas.

However, in the time series domain, datasets are not that easy to access, where there are usually a number of privacy issues, and it is typically difficult to obtain large enough or balanced datasets. This often leads to a major problem when attempting to train one of these models on an incomplete, unbalanced, or privacy-challenged dataset. Typically, these problems are addressed by pre-processing dataset techniques, such as subsampling, or in datasets that are not large enough, by \gls{da} techniques \cite{reviewDataAugmentation, imbalancedata}.

Nevertheless, as soon as problems arise, technology evolves to address these boundaries. In recent years, \glspl{ann} and their application to the field of \gls{dl} have experienced a period of great advances. Although a multitude of models has contributed to this expansion, one of the most revolutionary that has been proposed appeared in 2014 by Ian Goodfellow~\cite{goodfellow2014}, with his~\glspl{gan}.

\glspl{gan} are certainly not the earliest generative architecture ever introduced; already in 1987 Yann Lecun~\cite{autolecun} suggested in his thesis the \gls{ae} architectures, which were capable of generating data modifications received as input. But it is not until the incorporation of directed probabilistic models into \gls{ae} architectures, also known as \gls{vae}~\cite{vae2013}, that the models begin to be presented as capable of generating synthesised data.

Although these networks show impressive results, the capabilities of \glspl{gan} have been shown to be far ahead, with impressive results applied to the field of imaging. Furthermore, this is not the unique area of application. Synthetic data generation is a powerful boost for synthesis of sensitive data, such as in the world of telecommunications.

There are many time series applications where these algorithms have shown good results when improving the capabilities of the models by enlarging the datasets \cite{navidan2021generative, rigaki2018bringing}. In this area, time series data are especially sensitive due to aspects such as the low availability of high-quality datasets or privacy of the data \cite{mozo2022synthetic, wang2021deep, cao2022simulation, bhowmik2015spatially}.

All these advances in the \gls{dl} area allow us to synthetically increase the size of the datasets used in \gls{ml} tasks. As mentioned above, the size of the dataset is a sensible characteristic that can affect the performance of the developed models.

Regarding time series data, the availability of these huge datasets is even more complicated. There are other fields such as image processing or \gls{nlp} where data is much more available. But in the time series domain, it is more difficult to obtain these samples \cite{wang2021deep, cao2022simulation, bhowmik2015spatially}. In this sense, \gls{da} arises as a technique to counter the scarcity of data.

Another important factor is that time series data have particularities when it comes to processing them. Each dataset of time series is very different and needs special attention on how it is being augmented. Therefore, the proposed techniques for time series data must be analysed and discussed to better understand which technique can be used in which data.

Thus, this paper aims to review all the existing technologies for \gls{da}, and to review the positive and negative aspects of each of them. This review can help researchers better understand how \gls{da} techniques can be applied to time series data to obtain better results when training \gls{ml} models. We also expect that it will be useful to highlight the main differences between time series data and other domains.

\subsection{Paper structure}
The rest of the paper is structured as follows: Section \ref{section:ProblemStatement}: ``\nameref{section:ProblemStatement}'' presents the review in its context, providing a view of the importance of this work in the current context; Section \ref{section:RelatedWorks}: ``\nameref{section:RelatedWorks}'' reviews previous similar works in this area, highlighting the differences between the previous ones and the present manuscript; Section \ref{section:Background}: ``\nameref{section:Background}'' introduces a technical background of the performance of the techniques used to augment time series data; Section \ref{section:evaluationmetrics}: ``\nameref{section:evaluationmetrics}'' explains the problematic associated with how to evaluate the results of the new synthetic data; Section \ref{section:DAAlgorithms}: ``\nameref{section:DAAlgorithms}'' presents the most important and current work in this area, reviewing the technical aspects of each approximation; Section \ref{section:Discussion}: ``\nameref{section:Discussion}'' discusses the results and behavior of each algorithm presented in the previous section and finally Section \ref{section:Conclusion}: ``\nameref{section:Conclusion}'' summarizes the main conclusions of the research.

\section{Problem statement}
\label{section:ProblemStatement}
This study addresses the current context of generating new data samples in temporal series datasets. The purpose of this paper is to focus on the most relevant and cutting-edge research in this area to provide an accurate and complete view of this field. This research tries to cover all the possible techniques used to enhance this type of data.

In contrast to previous works, the aim of this survey is to provide a complete view of how different approaches to this problem have been developed. One of the main problems of reviewing a portion of the total techniques used in this area is that it does not fully explain how the current paradigm is structured. This is particularly critic when comparing different techniques, where it is desirable to have all the approximations in the area to fully understand the particularities of each one.

In this sense, it is important to provide an updated and comprehensive study of the state of the art in this area. This manuscript is focused on the three most important pillars of current data augmentation in time series: traditional algorithms \gls{vae} and \gls{gan}.

\section{Related works}
\label{section:RelatedWorks}
In the latest times, several high-quality data augmentation review papers have been published \cite{genericsurvey, surveyimages, nlpSurvey}. However, most of them are focused on more popular areas such as imaging, video or \gls{nlp}. Although these techniques focus on correcting the imbalance or incompleteness of the dataset, there are other areas of application where these problems are more common. The scarcity of valid datasets is not as clear in all areas of \gls{dl} applications as in time series. 

In a first approach to the literature review, in \cite{iwana2021empirical} an approximation of \acrshort{da} algorithms is made for use in neural network algorithms for time series classification. In the survey, they evaluated 12 methods to enhance time series data in 128 time series classification datasets with six different types of neural networks. Other recent studies focus more specifically on the use of GANs for data augmentation, as in \cite{brophy2021generative}, where they analyse the taxonomy of discrete-variant GANs and continuous-variant GANs, in which GANs deal with discrete time series and continuous time series data. These surveys analyse \gls{da} in time series using neural networks, but lack a comparison of these algorithms with more traditional approaches.

However, improving data sources to feed \gls{ai} algorithms is not limited to \gls{da} exclusively. Therefore, some studies have decided to take the path of building synthetic traffic generators to build their datasets almost from scratch; some examples focusing on this aspect are \cite{syntetictrafficcongress, syntheticdatageneration, syntheticdagaGAN}. In this way, they are able to abstract from the dataset itself, which is only necessary to understand the distribution of the data. Furthermore, in \cite{syntheticanalisis} they set out a further study of the repercussions of these technologies, highlighting one of the major advantages of generating synthesised data, the abstraction of privacy issues, and the ease of obtaining datasets.

Despite the possibilities presented by new technologies to improve the quality of time series datasets, there are no studies that compile all technologies, with a comprehensive comparison of them. Furthermore, to the authors' knowledge, a survey that compiles all the different techniques is missing in the literature. This work’s goal is precisely to address this problem. It is crucial to develop a review that studies and compares all novel techniques proposed in time series domain. Previous works were centred on a specific model or approximation, with works such as \cite{wen2020time, lim2021time, brophy2021generative}. The objective of this survey is to have a wider view of the field, and to be able to further compare each technology and different approximations. With respect to previous work, a complete review is presented, extending the explanation of each algorithm's performance, results, advantages and disadvantages.

In addition, this field is in constant evolution, so it is common for reviews to become outdated due to the publication of new articles. It is considered very important to provide an updated study of the state of the art that follows the latest trends in this area.

Therefore, the goal of this review is to contribute to reducing the existing gap in the area by trying to bring together all the time series \gls{da} algorithms that currently exist, contrasting their possible virtues, approaches, and differences to help future researchers position themselves in the area.

\section{Background}
\label{section:Background}
\subsection{Traditional algorithms}
\gls{da} has been a crucial task when the available data are unbalanced or insufficient. Traditionally, in fields such as image recognition, different transformations have been applied to data such as cropping \cite{krizhevsky2009learning_alexnet, simonyan2014very_vgg, he2016deep_resnet, szegedy2015going_inception}, scaling \cite{simonyan2014very_vgg, he2016deep_resnet}, mirroring \cite{krizhevsky2009learning_alexnet, huang2017densely_densenet, szegedy2015going_inception}, colour augmentation \cite{krizhevsky2009learning_alexnet, he2016deep_resnet, mikolajczyk2018data} or translation \cite{huang2017densely_densenet}.

These algorithms cannot be applied directly to time series given the particularity of the time series data distribution \cite{lashgari2020data}. E.g., if one wants to apply rotation to augment an image dataset, it is possible to rotate each image to generate new ones. This cannot be directly done in the time series domain, e.g., if a time series sample is divided into several portions and these portions are reorganised using linear interpolation between them, the result would not be valid because the tendency of the data would be destroyed. Due to the diversity of the time series data, not all techniques can be applied to every dataset. Some of the previous algorithms used in computer vision must adapted to a time series domain, but, in other cases, new specific algorithms must be designed to treat with time series data.

Another important factor when applying \gls{da} to the time series domain, especially in signal processing, is that manipulation of the data could distort the signal too much, leading to negative training.

Traditional algorithms are defined as all the techniques based on taking data input samples and synthesising new samples by modifying these data and applying different transformations. The main difference between this technique and those that are reviewed in Sections \ref{section:VAE} and \ref{section:GAN} is that, in the former algorithms, the transformations are applied directly to the data, while in the latter the objective is to learn the probability distribution of the data in order to generate completely new samples trying to imitate the data distribution.

\subsection{\acrfull{vae}}
\label{section:VAE}
\gls{vae}s are neural generative models first introduced by Diederik P. Kingma and Max Welling\cite{vae2013}. This algorithm is based on the \gls{ae} architecture\cite{autolecun} proposed in 1987. \gls{ae}s allow changing typical artificial intelligence problems, such as linear regression or classification, to domain-shifting problems. In order to perform this, \gls{ae}s take an input, usually an image, and infer as the output modifications of that same input. This is known as self-supervised training, where the objective is to obtain the input with slight modifications as an output. One of the most popular applications of this model is image denoising \cite{denoising}. In this case, the input is an image that contains noise and the output should be the input image without the undesired noise.

\gls{ae} Network is composed of two components, an Encoder and a Decoder. The Encoder is in charge of reducing the input dimensionality of the data to a latent space, while the Decoder reconstructs the input information from this latent representation. This latent space is a lower-dimensional manifold of the input data. Then, synthetic data are generated, interpolating the values of the latent space and decoding them. However, this interpolation of the latent space does not generate completely new values; it just mixes the features of the learned probability distribution.

In order to avoid the overfitting produced in \gls{ae}, \gls{vae} regularises its training, generating more diverse samples. The main difference between both architectures is that \gls{vae} encodes the input information in a probability distribution rather than in a point. Then, from this distribution, it samples a point that is then decoded to synthesise new samples.

This intermediate step allows the network to map the input distribution to a lower-dimensional distribution from which new latent points can be generated. To do so, the latent distribution is normally defined by a normal distribution with a mean $ \vec{\mu}=\left(\mu_{1}, \ldots, \mu_{n}\right) $ and a standard deviation $ \vec{\sigma}=\left(\sigma_{1}, \ldots, \sigma_{n}\right) $. These mean and standard deviation vectors define the latent distribution of the model.

Leaving the network to learn a distribution, instead of a set of points learned in \gls{ae}, the decoder network associates the features of the input data with the probability areas with their respective mean and deviation. With this representation, the mean of the distribution defines the center point from which the synthetic samples will be generated and the standard deviation defines the variability in the output, that is, the diversity of the generated samples.

\begin{figure}[h]
	\centering
	\includegraphics[width=.6\textwidth]{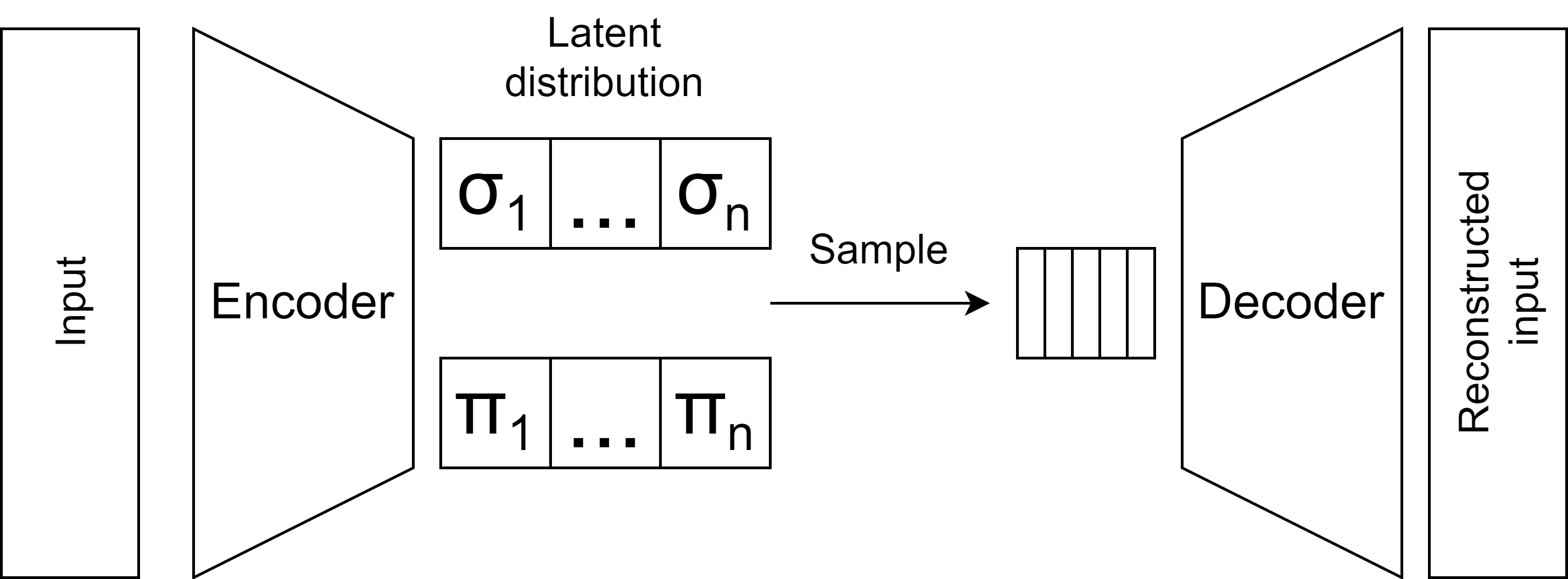}
    \caption{\gls{vae} architecture with the both latent space representations ($\pi$ and $\sigma$).}
    \label{figure:AEVAEArchitecture}
\end{figure}

Figure \ref{figure:AEVAEArchitecture} shows the architecture of a \gls{vae} network.

Regarding the training of a \gls{vae} network there are two different loss functions. The \textit{reconstruction term} is in charge of the reconstruction of the input data. It measures the error of the network when it builds. This metric acts the same as the error of a standard \gls{ae}.

On the other hand, \glspl{vae} include a \textit{regularisation term} that tries to organise how the latent distribution generates new latent spaces. The function of this term is to measure the distance of the sampled data points and a Gaussian distribution. The distance used to measure this error is the \gls{kldivergence} \cite{kullback1951information}. So in order to calculate the loss of the \gls{vae} both errors are added, measuring at the same time the reconstruction error of the output and the error of sampling points following a Normal distribution.

\glsreset{gan}
\subsection{\glspl{gan}}
\label{section:GAN}
\glspl{gan} are a generative neural model based on a competition between two \glspl{nn}. They were first introduced by Ian Goodfellow\cite{goodfellow2014} in 2014. The objective of the architecture is to replicate a given data distribution in order to synthesise new samples of the distribution. To achieve this goal, the \gls{gan} architecture is composed of a \gls{g} model and a \gls{d} model. The former is in charge of generating the synthetic samples of the data distribution, while the latter tries to distinguish the real samples from the synthesised samples.

To accomplish such generation of completely new data that are indistinguishable from the input data distribution, both models interact with each other. The model \gls{g} generates samples trying to replicate, without copying, the distribution, while the model \gls{d} discriminates the real samples from the fake samples. In this way, when \gls{d} differentiates both distributions, it feedbacks \gls{g} negatively; on the other hand, when \gls{d} is not capable of differentiating each distribution, its positively feedbacks \gls{g}. In doing so, \gls{g} evolves to fool \gls{d}. At the same time, \gls{d} is positively rewarded when discrimination is done correctly.

\begin{figure}[h]
    \centering
    \includegraphics[width=.7
    \textwidth]{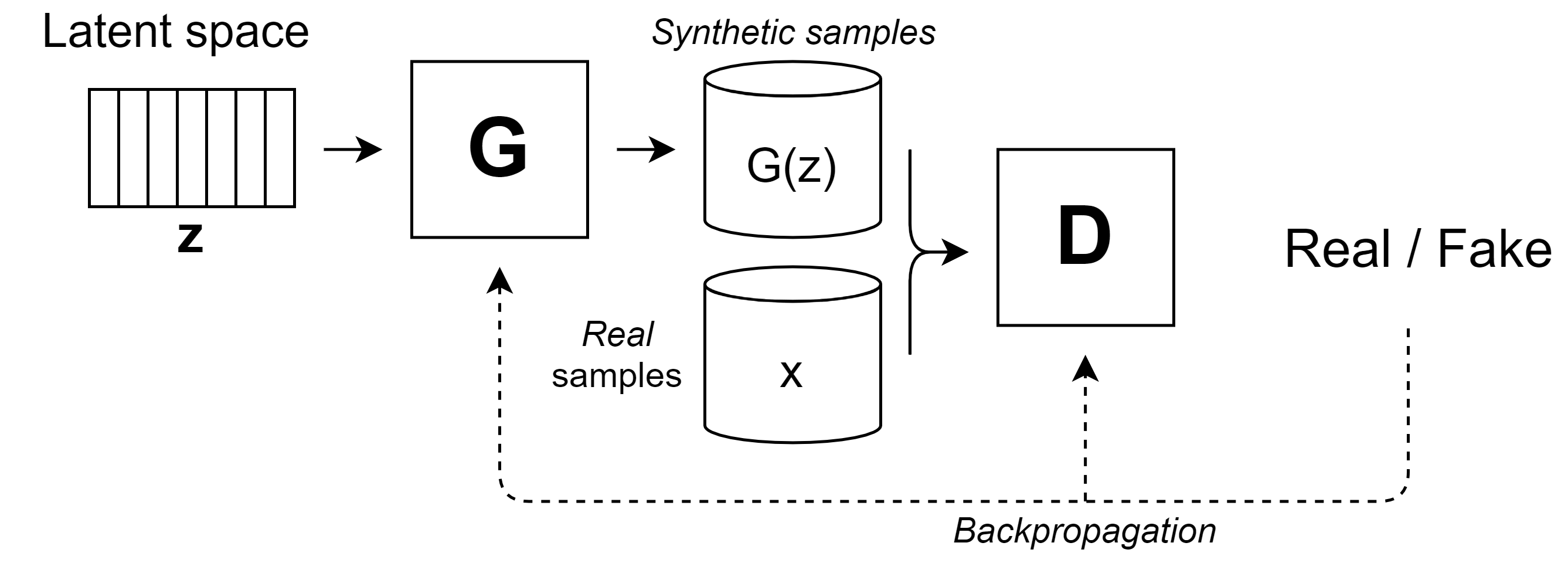}
    \caption{\gls{gan} architecture.}
    \label{figure:GANArchitecture}
\end{figure}

This competition encourages both networks to evolve together. If \gls{d} fails in its task, \gls{g} will not evolve because it will always succeed, despite the quality of the synthesised samples. Even if \gls{d} always perfectly distinguishes both distributions, \gls{g} will not be able to fool \gls{d}, making it impossible to evolve.
The standard \gls{gan} architecture is depicted in Figure \ref{figure:GANArchitecture}.

From a mathematical perspective, this competitive behavior is based on Game Theory, where two players compete in a zero-sum game. The \gls{d} estimates the \emph{a posteriori} probability $ p(y|x) $, where $ y $ is the label (true or fake) of the given sample $ x $. And \gls{g} generates synthetic samples from a latent vector $ z $, which can be denoted as $ G(z) $.

From a formal point of view, this competition is defined as a minimax game where \gls{d} tries to maximize its accuracy when discriminating between both distributions and \gls{g} tries to minimize this accuracy. The formulation of this process is denoted as follows:
\begin{equation}
    \begin{array}{r}
    \min _{G} \max _{D} L(D, G) = \min _{G} \max _{D} E_{x \sim p_{r}} \log [D(x)] +E_{z \sim p_{z}} \log [1-D(G(x))],
    \end{array}
\end{equation}
where $ z $ is the latent vector, which is generated randomly by a uniform or Gaussian distribution $p_z$ and $ x \sim p_{r} $ is the real distribution.

In the publication in which \glspl{gan} was presented\cite{goodfellow2014} it was proved that the architecture can converge to a unique solution. This point, known as Nash equilibrium, is characterized by the fact that none of the networks can reduce their respective losses. This optimal result is very difficult to achieve in reality due to the instability behavior of \gls{gan}. The Nash equilibrium is, in fact, most of the time not achieved because of the constant competition between both networks.

Respecting \gls{da}, \glspl{gan} synthesise completely new samples, using the dataset distribution as their base on learning the underlying data distribution. Therefore, \glspl{gan} are able to produce more diverse samples respect with respect to previous approximations. This is known as data generation, where the main difference between data generation and \gls{da} is that the former generates new samples of synthetic data while the latter use the original samples to produce new ones using their information.

\section{Evaluation metrics}
\label{section:evaluationmetrics}
Due to the particularities of the time series field, there is no unique metric to evaluate the reliability of algorithms in all of their applications. Finding a measurement capable of evaluating the quality and diversity of the synthesised data is still an open issue.

For example, in \gls{gan} networks, there exists no consensus between the different studies about the evaluation metrics to use. In addition, most of the evaluation metrics designed are centred on computer vision, since it is the most popular field for this kind of network.

Therefore, it will be described the most commonly used metrics that have been used to evaluate the algorithms that will be discussed in this article. However, it should be noted that to choose a proper evaluation metric, one should adapt the metric to the specific data augmentation algorithm and the application field.

\subsection{External performance evaluation}
When applying \gls{da} to a dataset, the most common objective is to generate new data samples to improve the performance of certain models, reducing the imbalance of the data or the lack of data. One of the most popular ways to measure how the addition of new data changes the behavior of the models is simply to compare these models before and after \gls{da}. Then it is possible to compare whether each model has improved its performance after applying \gls{da} to the input data.

This approximation is purely practical and relies on the correlation between the performance of a defined model and not the quality of the synthetic samples themselves. Most traditional algorithms base their performance on this method because it is a straightforward method to evaluate an algorithm.

In \cite{bandara2021improving} the performance of the \gls{da} algorithms they propose is achieved using \textit{symmetric Mean Absolute Percentage Error} and \textit{Mean Absolute Scaled Error}, which are the two most common evaluation metrics used in forecasting. This research compares the values of these metrics before and after applying \gls{da} to the dataset, then evaluates how the models improve their performance due to the addition of more data to the training set.

In \cite{iwana2021empirical} they used six different neural networks to evaluate how each \gls{da} algorithm affects the classification of the data. In particular, they evaluated VGG\cite{simonyan2014very_vgg}, Residual Network\cite{he2016deep_resnet}, Multilayer Perceptron\cite{wang2017time_mlp}, Long Short-Term Memory\cite{reimers2017optimal_lstm}, Bidirectional Long Short-Term Memory\cite{schuster1997bidirectional_blstm} and Long Short-Term Memory Fully Convolutional Network\cite{karim2017lstm_lstmfcn}. Then, the changes in the accuracy of the models are compared, observing how certain \gls{da} algorithms benefit the performance of the models, while in other cases it gets worse. The main drawback stated in the article is that each architecture has its particularities, given the different results for each algorithm and making it a difficult task to differentiate the best algorithm. Furthermore, because all of them are neural models, it is difficult to interpret some of the results.

The approach followed in \cite{lashgari2020data} is to compare different \gls{da} techniques by the increase in accuracy produced in each case of study. It is worth mentioning how this approximation adapts to each application without having to change anything. The authors of the article are able to compare very different techniques such as Noise Addition, \gls{gan}, Sliding Window, Fourier-Transform and Recombination of Segmentation under the same criteria for a specific domain purpose. This example shows how this approach easily adapts to different \gls{da} techniques, making it possible to compare the results for a certain task.

A similar strategy to measure the quality of the generated data is to compare different models using a defined loss function. This approach was followed in \glspl{gan} architectures in works like \cite{goodfellow2014, isola2017image_pix2pix, yi2017dualgan, zhu2017unpaired_cyclegan, wang2018high_ps2man}, where a comparison between networks is possible using the same loss function to evaluate their training. Then, they correlate the quality of the synthetic data with this value. This strategy can be applied naturally to the time series domain, allowing comparison between different networks. However, the main drawback of this method is that it compares the performance of different neural models and cannot be applied to other models. It should be noted that, as in previous metrics, it correlates the quality of the generated data not with the data itself but rather with the performance of the model.

In \cite{yang2021voice2series} they compare the performance of different \gls{da} techniques with \gls{mpce}. This metric, proposed in \cite{wang2017time}, measures the error per class in $J$ datasets taking into account the number of classes in each dataset. The main particularity of \gls{mpce} is that it allows us to quantify the performance of an algorithm for different datasets. The \gls{mpce} is calculated as follows:
\begin{equation}
    \text { MPCE }=\sum_{j \in[J]} \mathrm{PCE}_{j}=\frac{e_{j}}{c_{j}}
\end{equation}
where $ e_{j} $ is the error rate and $ c_{j} $ is the number of classes in each dataset. This metric is capable of taking into account the number of classes of each dataset, in order to normalise the comparison between different sources of data.

\subsection{\gls{gan} related metrics}
Since the introduction of \gls{gan} it has always been an open issue to measure the quality of the synthesised samples produced by the architecture\cite{borji2019pros}. One of the most important difficulties when designing a metric for \glspl{gan} is the ability to capture both the quality and diversity of the generated data.

In addition to being still an open issue, there is consensus on some metrics and many papers measure their results with the same metrics\cite{radford2015unsupervised_dcgan, karras2017progressive_progan, karras2019style_stylegan, karras2021alias_aliasfree, zhu2019dm_dmgan, gao2021lightweight_ldcgan}. The main problem in the time series domain is that it is not always possible to adapt the metrics to the particularities of this field because most of the metrics are designed to be useful in computer vision-related tasks.

Over the past few years, some works have suggested applying \gls{da} to time series data and treating it as if it were an image\cite{hartmann2018eeg, wang2020part}. These papers use \gls{gan} networks to synthesise new time series data, but to convert the signal data into an image. In these cases, traditional \gls{gan} metrics, such as the Inception Score\cite{salimans2016improved}, Mode Score\cite{gurumurthy2017deligan} or Fréchet Inception Distance\cite{heusel2017gans} are used to evaluate the results. These metrics are based on how the \textit{Inception v3} neural classifier distinguishes the different samples. The idea is to measure the entropy of the synthetic dataset using an external classifier.

In addition to the field of computer vision, studies have been developed that apply \gls{gan} directly to time series. That is, in \gls{timegan}\cite{yoon2019time_timegan} two new metrics are proposed to assess the quality of the generated samples. The Discriminative Score is based on the use of an external pre-trained model, as was done with the Inception Score, consisting of a 2-layer \gls{lstm}. The Discriminative Score measures how this model distinguishes between real and fake samples and the classification error corresponds to the Discriminative Score. The Predictive Score measurement was introduced in \cite{esteban2017real_rcgan} with the name \gls{tstr} which also uses a 2-layer \gls{lstm}, but in this case, this model is trained with synthetic samples. The model is then evaluated using the original dataset. The Predictive Score corresponds to \gls{mae} of the model trained with the synthetic samples evaluated with the real samples. This metric is, at the moment, one of the most effective and used evaluation metrics.

\subsection{Similarity measurements}
This set of metrics focuses on the comparison of two probability distributions. The idea is to measure how far from the original distribution the synthetic samples generated with \gls{da}. The main advantage of these metrics is that they focus on directly studying the quality of the data, in contrast the previously reviewed methods that measured the quality indirectly. Another advantage of these types of metrics is that they can be applied to synthetic data despite the algorithm used to generate them.

An empirical and qualitative approach to measuring the differences between two distributions is to reduce the dimensionality of the data and perform a visual comparison. The objective is to reduce the dimensionality of the data to plot the samples in a bidimensional space; an empirical comparison is then made by visualising the data. This approach was followed in \cite{naaz2021generative} where they applied \gls{tsne} and \gls{pca}. Then they compared the distribution of the data in the two-dimensional space for TimeGAN\cite{yoon2019time_timegan}, \gls{rcgan}\cite{esteban2017real_rcgan}, \gls{crnngan}\cite{mogren2016c-crnngan}, T-Forcing\cite{sutskever2011generating_fforcing}, WaveNet\cite{van2016wavenet} and WaveGAN\cite{donahue2018adversarial_wavegan}. This approach was also followed in \cite{iwana2021empirical} where they used \gls{pca} to compare different traditional algorithms for the GunPoint dataset from the 2018 UCR Time Series Archive\cite{UCRArchive2018}.

 \gls{kldivergence} has been used in work such as \cite{fu2020data, debnath2021exploring} to measure similarities between synthetic and real datasets. Recall that \gls{kldivergence} \cite{kullback1951information} is defined as
\begin{equation}
    D_{K L}(P \,||\, Q)=\sum_{i} P(x_i) \log\left(\frac{P(x_i)}{Q(x_i)}\right),
\end{equation}
where $ P $ and $ Q $ are the probability distributions whose distance is calculated and $i$ runs over the samples $x_i$ of the distribution. This \gls{kldivergence} \cite{kullback1951information} is not a symmetric distance, so it can be symmetrized to give rise to the so-called \gls{JSD}, defined as
$$
    \textrm{JSD}(P \,||\, Q) = D_{K L}(P \,||\, (P+Q)/2) + D_{K L}(Q \,||\, (P+Q)/2).
$$

In \cite{cazelles2020wasserstein} a novel measurement is proposed to quantify the distance between the time series distribution. It is based on calculating the Wasserstein distance between time series data. The metric is defined by measuring the Wasserstein distance of the energy between frequencies. The Wasserstein-Fourier distance between the probability distributions is computed as follows:
\begin{equation}
    \mathrm{WF}([x],[y])=W_{2}\left(s_{x}, s_{y}\right)
\end{equation}
where $ s_{x} $ and $ s_{y} $ are the normalised power spectral densities of the distributions.

\section{Data Augmentation algorithms review}
\label{section:DAAlgorithms}
During this section, different state-of-the-art algorithms will be reviewed. This section will explain the particularities, strengths and weaknesses of each algorithm. In addition, the different approximations to apply \gls{da} will be grouped and related between them. A taxonomy of the different trends and lines of research will be proposed, showing the different existing links between the works of the last years.

It should be noted that not all the algorithms can be applied to all types of time series data; in some cases, the algorithms proposed will be heavily focused on a certain application, while in others more general techniques will be studied.

\subsection{Basic \gls{da} Methods}
The basic \gls{da} algorithms that will be reviewed in this section are all techniques that use data manipulation to generate new synthetic data samples using existing samples and transform the original samples. All these techniques have as their base the deformation, shortening, enlargement or modification of the data samples of the dataset. This group of techniques has been traditionally used in fields such as computer vision and, in some cases, the same algorithms can be adapted to process time series data, but in others, new algorithms must be designed specifically to use time series data as input.

Therefore, the most important traditional algorithms that have been applied to time series data will be reviewed and discussed, outlining their particularities, advantages and disadvantages. Figure \ref{figure:TraditionalAlgorithms_Taxonomy} shows the taxonomy proposed for the different algorithms reviewed.

\tikzset{every picture/.style={line width=0.75pt}} 
\begin{figure}
\begin{tikzpicture}[x=0.75pt,y=0.75pt,yscale=-1,xscale=1]
\draw    (183.5,270.25) -- (250.73,432.4) ;
\draw [shift={(251.5,434.25)}, rotate = 247.48] [color={rgb, 255:red, 0; green, 0; blue, 0 }  ][line width=0.75]    (10.93,-3.29) .. controls (6.95,-1.4) and (3.31,-0.3) .. (0,0) .. controls (3.31,0.3) and (6.95,1.4) .. (10.93,3.29)   ;
\draw    (183.5,270.25) -- (250.28,357.16) ;
\draw [shift={(251.5,358.75)}, rotate = 232.46] [color={rgb, 255:red, 0; green, 0; blue, 0 }  ][line width=0.75]    (10.93,-3.29) .. controls (6.95,-1.4) and (3.31,-0.3) .. (0,0) .. controls (3.31,0.3) and (6.95,1.4) .. (10.93,3.29)   ;
\draw    (183.5,270.25) -- (249.54,283.36) ;
\draw [shift={(251.5,283.75)}, rotate = 191.23] [color={rgb, 255:red, 0; green, 0; blue, 0 }  ][line width=0.75]    (10.93,-3.29) .. controls (6.95,-1.4) and (3.31,-0.3) .. (0,0) .. controls (3.31,0.3) and (6.95,1.4) .. (10.93,3.29)   ;
\draw    (183.5,270.25) -- (250.02,209.6) ;
\draw [shift={(251.5,208.25)}, rotate = 137.64] [color={rgb, 255:red, 0; green, 0; blue, 0 }  ][line width=0.75]    (10.93,-3.29) .. controls (6.95,-1.4) and (3.31,-0.3) .. (0,0) .. controls (3.31,0.3) and (6.95,1.4) .. (10.93,3.29)   ;
\draw    (183.5,270.25) -- (249.63,133.55) ;
\draw [shift={(250.5,131.75)}, rotate = 115.82] [color={rgb, 255:red, 0; green, 0; blue, 0 }  ][line width=0.75]    (10.93,-3.29) .. controls (6.95,-1.4) and (3.31,-0.3) .. (0,0) .. controls (3.31,0.3) and (6.95,1.4) .. (10.93,3.29)   ;
\draw    (183.5,270.25) -- (250.4,58.66) ;
\draw [shift={(251,56.75)}, rotate = 107.54] [color={rgb, 255:red, 0; green, 0; blue, 0 }  ][line width=0.75]    (10.93,-3.29) .. controls (6.95,-1.4) and (3.31,-0.3) .. (0,0) .. controls (3.31,0.3) and (6.95,1.4) .. (10.93,3.29)   ;
\draw    (304.5,57.75) -- (383.64,89.51) ;
\draw [shift={(385.5,90.25)}, rotate = 201.86] [color={rgb, 255:red, 0; green, 0; blue, 0 }  ][line width=0.75]    (10.93,-3.29) .. controls (6.95,-1.4) and (3.31,-0.3) .. (0,0) .. controls (3.31,0.3) and (6.95,1.4) .. (10.93,3.29)   ;
\draw    (304.5,57.75) -- (384.51,49.94) ;
\draw [shift={(386.5,49.75)}, rotate = 174.43] [color={rgb, 255:red, 0; green, 0; blue, 0 }  ][line width=0.75]    (10.93,-3.29) .. controls (6.95,-1.4) and (3.31,-0.3) .. (0,0) .. controls (3.31,0.3) and (6.95,1.4) .. (10.93,3.29)   ;
\draw    (309.5,208.25) -- (384.93,148.99) ;
\draw [shift={(386.5,147.75)}, rotate = 141.84] [color={rgb, 255:red, 0; green, 0; blue, 0 }  ][line width=0.75]    (10.93,-3.29) .. controls (6.95,-1.4) and (3.31,-0.3) .. (0,0) .. controls (3.31,0.3) and (6.95,1.4) .. (10.93,3.29)   ;
\draw    (309.5,208.25) -- (385,208.74) ;
\draw [shift={(387,208.75)}, rotate = 180.37] [color={rgb, 255:red, 0; green, 0; blue, 0 }  ][line width=0.75]    (10.93,-3.29) .. controls (6.95,-1.4) and (3.31,-0.3) .. (0,0) .. controls (3.31,0.3) and (6.95,1.4) .. (10.93,3.29)   ;
\draw    (309.5,208.25) -- (384.63,179.95) ;
\draw [shift={(386.5,179.25)}, rotate = 159.36] [color={rgb, 255:red, 0; green, 0; blue, 0 }  ][line width=0.75]    (10.93,-3.29) .. controls (6.95,-1.4) and (3.31,-0.3) .. (0,0) .. controls (3.31,0.3) and (6.95,1.4) .. (10.93,3.29)   ;
\draw    (309.5,208.25) -- (384.63,237.03) ;
\draw [shift={(386.5,237.75)}, rotate = 200.96] [color={rgb, 255:red, 0; green, 0; blue, 0 }  ][line width=0.75]    (10.93,-3.29) .. controls (6.95,-1.4) and (3.31,-0.3) .. (0,0) .. controls (3.31,0.3) and (6.95,1.4) .. (10.93,3.29)   ;
\draw    (309.5,208.25) -- (384.91,266.03) ;
\draw [shift={(386.5,267.25)}, rotate = 217.46] [color={rgb, 255:red, 0; green, 0; blue, 0 }  ][line width=0.75]    (10.93,-3.29) .. controls (6.95,-1.4) and (3.31,-0.3) .. (0,0) .. controls (3.31,0.3) and (6.95,1.4) .. (10.93,3.29)   ;

\draw    (106,248) -- (183,248) -- (183,294) -- (106,294) -- cycle  ;
\draw (109,252) node [anchor=north west][inner sep=0.75pt]   [align=left] {Traditional\\algorithms};
\draw    (252,45) -- (304,45) -- (304,70) -- (252,70) -- cycle  ;
\draw (255,49) node [anchor=north west][inner sep=0.75pt]   [align=left] {Slicing};
\draw  [color={rgb, 255:red, 74; green, 144; blue, 226 }  ,draw opacity=1 ]  (252,120.2) -- (313,120.2) -- (313,145.2) -- (252,145.2) -- cycle  ;
\draw (255,124.2) node [anchor=north west][inner sep=0.75pt]   [align=left] {Jittering};
\draw    (252,195.45) -- (309,195.45) -- (309,220.45) -- (252,220.45) -- cycle  ;
\draw (255,199.45) node [anchor=north west][inner sep=0.75pt]   [align=left] {Scaling};
\draw  [color={rgb, 255:red, 74; green, 144; blue, 226 }  ,draw opacity=1 ]  (252,270.6) -- (316,270.6) -- (316,295.6) -- (252,295.6) -- cycle  ;
\draw (255,274.6) node [anchor=north west][inner sep=0.75pt]   [align=left] {Rotation};
\draw  [color={rgb, 255:red, 74; green, 144; blue, 226 }  ,draw opacity=1 ]  (252,345.8) -- (341,345.8) -- (341,370.8) -- (252,370.8) -- cycle  ;
\draw (255,349.8) node [anchor=north west][inner sep=0.75pt]   [align=left] {Permutation};
\draw  [color={rgb, 255:red, 74; green, 144; blue, 226 }  ,draw opacity=1 ]  (252,421) -- (399,421) -- (399,446) -- (252,446) -- cycle  ;
\draw (255,425) node [anchor=north west][inner sep=0.75pt]   [align=left] {Channel permutation};
\draw  [color={rgb, 255:red, 74; green, 144; blue, 226 }  ,draw opacity=1 ]  (387,27) -- (521,27) -- (521,73) -- (387,73) -- cycle  ;
\draw (390,31) node [anchor=north west][inner sep=0.75pt]   [align=left] {Concatenating and\\resampling};
\draw  [color={rgb, 255:red, 74; green, 144; blue, 226 }  ,draw opacity=1 ]  (387,76.33) -- (528,76.33) -- (528,101.33) -- (387,101.33) -- cycle  ;
\draw (390,80.33) node [anchor=north west][inner sep=0.75pt]   [align=left] {Time slicing window};
\draw  [color={rgb, 255:red, 74; green, 144; blue, 226 }  ,draw opacity=1 ]  (387,135) -- (543,135) -- (543,160) -- (387,160) -- cycle  ;
\draw (390,139) node [anchor=north west][inner sep=0.75pt]   [align=left] {Homogeneous scaling};
\draw  [color={rgb, 255:red, 74; green, 144; blue, 226 }  ,draw opacity=1 ]  (387,165.25) -- (522,165.25) -- (522,190.25) -- (387,190.25) -- cycle  ;
\draw (390,169.25) node [anchor=north west][inner sep=0.75pt]   [align=left] {Magnitude warping};
\draw  [color={rgb, 255:red, 74; green, 144; blue, 226 }  ,draw opacity=1 ]  (387,195.5) -- (523,195.5) -- (523,220.5) -- (387,220.5) -- cycle  ;
\draw (390,199.5) node [anchor=north west][inner sep=0.75pt]   [align=left] {Frequency warping};
\draw  [color={rgb, 255:red, 74; green, 144; blue, 226 }  ,draw opacity=1 ]  (387,225.75) -- (484,225.75) -- (484,250.75) -- (387,250.75) -- cycle  ;
\draw (390,229.75) node [anchor=north west][inner sep=0.75pt]   [align=left] {Time warping};
\draw  [color={rgb, 255:red, 74; green, 144; blue, 226 }  ,draw opacity=1 ]  (387,256) -- (505,256) -- (505,281) -- (387,281) -- cycle  ;
\draw (390,260) node [anchor=north west][inner sep=0.75pt]   [align=left] {Window warping};
\end{tikzpicture}

\caption{Traditional \gls{da} algorithms taxonomy.}
\label{figure:TraditionalAlgorithms_Taxonomy}
\end{figure}
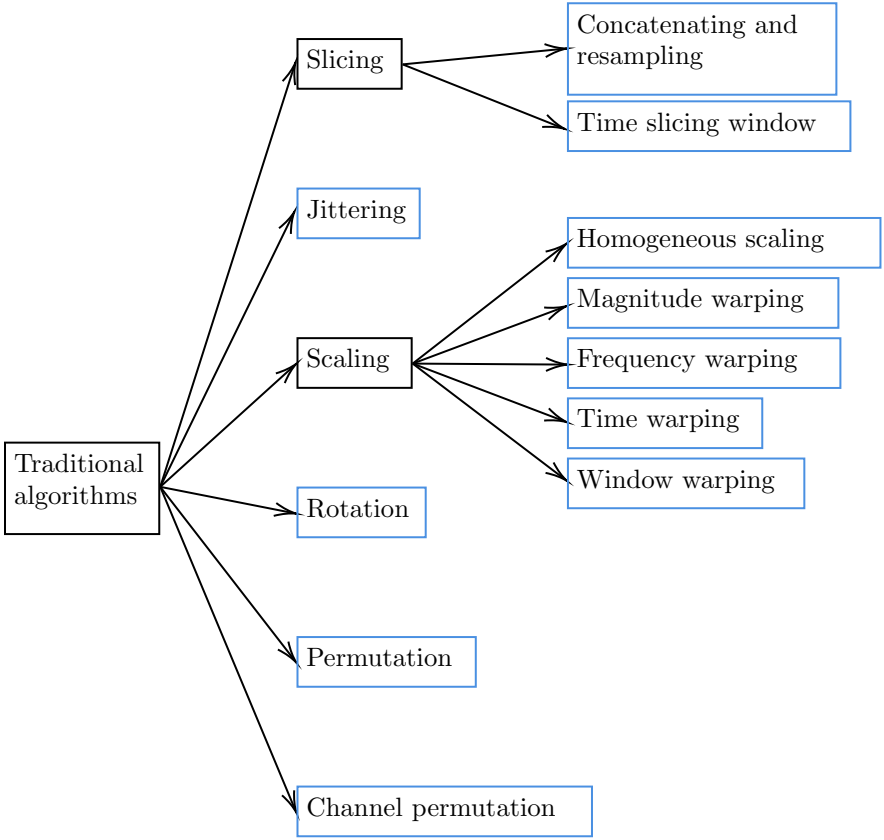

\subsubsection{Time slicing window}
Slicing, in time series, consists of cutting a portion of each data sample, to generate a different new sample. Normally, slicing is applied to the last steps of the sample, but the snippet of the original sample can be obtained from any step. When the original data is cropped, a different sample is produced, but unlike image processing, it is difficult to maintain all the features of the original data. The process of slicing time series data provides new data given as:
\begin{equation}
    x^{\prime}(W)=\left\{x_{\varphi}, \ldots, x_{t}, \ldots, x_{\varphi+W}\right\},
\end{equation}
where $ W $ is the slice window that defines the crop size and $ \varphi $ is the initial point from where the slicing is performed, such as $ 1 \leq \varphi \leq T-W $. One of the most important drawbacks of slicing the signal is that it can lead to invalid synthetic samples because it can cut off important features of the data.

A variation of the slicing method is proposed in \cite{cao2020novel}, where the concatenating and resampling method is presented. This algorithm first detects features in the data, called characteristic points. This is made by using the Pan-Tompkins QRS detector\cite{pan1985real}. This algorithm detects the characteristic points in a heartbeat signal, so in order to apply the concatenating and resampling algorithm it must be defined and algorithm to detect these points. Then, after detecting the characteristic points, it is defined a subsequence that starts and ends in a characteristic point. This sequence is replicated several times and sliced in a window to perform \gls{da}.

This variation was applied to \gls{ecg} data of variable length between 9 and 61 seconds sampled at 300 Hz.

The concatenating and resampling algorithm tries to ensure the validity of the data, taking into account that the signal maintains its features. But the main disadvantage of this method is that it needs a detector of characteristic points that ensure the data validity.

\subsubsection{Jittering}
Jittering consists of adding noise to time series to perform \gls{da}. This technique, in addition to being one of the simplest forms of \gls{da} is one of the most popular in time series\cite{flores2021data, rashid2019window}. Jittering assumes that the data are noisy which, in many cases, i.e., when dealing with sensor data, is true.

Jittering tries to take advantage of the noise of the data and simulate it to generate new samples. Typically, Gaussian noise is added to each time step; the mean and standard deviation of this noise define the magnitude and shape of the deformation, so it is different in each application. The jittering process can be defined as follows:
\begin{equation}
    x^{\prime}(\epsilon)=\left\{x_{1}+\epsilon_{1}, \ldots, x_{t}+\epsilon_{t}, \ldots, x_{T}+\epsilon_{T}\right\},
\end{equation}
where $ \epsilon $ stands for the noise addition at each step of the signal.

As mentioned above, the jittering process must be adapted to each case, because there are cases such as \cite{um2017data} where the effects of jittering lead to negative learning. In this research, it was used as time series data the information received by a wearable sensor, capturing 58 seconds at 62.5 Hz that were later resampled to 120 Hz per sample.

\subsubsection{Scaling}
Scaling consists of changing the magnitude of a certain step in the time series domain. The idea is to maintain the overall shape of the signal while changing its values. With scaling, the new generated data change the range of values, but keep the shape of the changes. Homogeneous scaling is given as:
\begin{equation}
    x^{\prime}(\alpha)=\left\{\alpha x_{1}, \cdots, \alpha x_{t}, \cdots, \alpha x_{T}\right\},
\end{equation}
where $ \alpha > 0 $ defines the scale of the change. This value can be defined by a Gaussian distribution with mean 1 and with $ \sigma $ as a hyperparameter\cite{um2017data}, or it can be previously defined from a list of values\cite{rashid2019window}. 

Within scaling techniques, there are several different approximations for a specific time series domain. They take advantage of the specific properties of the signal data and adapt to perform \gls{da}.

Magnitude warping is a technique used in \cite{um2017data} that consists of an application of a variable scaling to different points of the data curve. To define where to apply the transformation, a set of knots $ \mathbf{u}=u_{1}, \ldots, u_{i}$ is defined; these represent the step in which the scaling is performed and their values are generated by using a normal distribution. Then the magnitude of the scaling is defined by a cubic spline interpolation of the knots $ S(\mathbf{x}) $. Then the magnitude warping can be defined as follows:
\begin{equation}
    x^{\prime}(\mathbf{\alpha})=\left\{\alpha_{1} x_{1}, \ldots, \alpha_{t} x_{t}, \ldots, \alpha_{T} x_{T}\right\},
\end{equation}
where $ \mathbf{\alpha}=\alpha_{1}, \ldots, \alpha_{i} = S(\mathbf{x}) $. With magnitude warping, the main particularity is that it applies a smoothened scaling to each point of the curve, multiplying the possibilities of the transformation while preserving the overall shape of the data. However, it still assumes that the synthetic data maintain validity after transformation.

This technique was used in \cite{um2017data} where the data was captured from a wearable device to detect if a patient suffers from Parkinson's disease.

Frequency warping is a variation of magnitude warping, mostly applied in speech processing\cite{adachi2007vocal, cui2015data, ko2015audio}. The most popular version in speech recognition is Vocal Tract Length Perturbation, which can be applied in a deterministic way\cite{cui2015data} or stochastically within a range\cite{jaitly2013vocal}. In particular, this technique was used in \cite{cui2015data} where a dataset of human conversation sampled at 8 KHz was used.

Another scaling technique is Time warping, the idea is very similar to magnitude warping, but the main difference between both algorithms is that Time warping modifies the curve in the temporal dimension. That is, instead of fluctuating the magnitude of the signal in each step, it stretches and shortens the time slices of the signal. To define how to warp the signal, a smooth curve, as was done in magnitude warping, is defined by using a cubic spline for a set of knots. The time-warping algorithm can be denoted as:
\begin{equation}
    x^{\prime}(\mathbf{\tau})=\left\{x_{\tau(1)}, \ldots, x_{\tau(t)}, \ldots, x_{\tau(T)}\right\},
\end{equation}
where $\tau $ defines the magnitude of the warp, this function is generated by using a cubic spline $ S(u) $ between different knots generated using a normal distribution. This algorithm has been used in several works, such as \cite{park2019specaugment, jeong2021sensor}. There is yet another variation of this algorithm, known as window warping, followed in \cite{le2016data} that defines a slice in the time series data and speeds up or down the data by a factor of $ 1/2 $ or $ 2 $. In this case, the warping is applied to a defined slice of the whole sequence; the rest of the signal is not changed.

\subsubsection{Rotation}
Rotation can be applied to multivariate time series data by applying a rotation matrix with a defined angle. In univariate time series, rotation can be applied by flipping the data. Rotation is defined as follows:
\begin{equation}
    x^{\prime}(R)=\left\{Rx_{1}, \ldots, Rx_{t}, \ldots, Rx_{T}\right\},
\end{equation}
where $ R $ is the rotation matrix used to twist the data. This algorithm is not very usual in time series due to the fact that rotating a time series sample could make it lose the class information, as it happened in \cite{fawaz2018data}, where it was used the UCR archive \cite{UCRArchive2018} with various dataset from different real-world applications were the composition of the samples of each dataset varies. On the other hand, there have been articles\cite{um2017data} that demonstrate the benefits of applying rotation, especially combined with other data transformations, in this case, using wearable data samples at 120 Hz.

\subsubsection{Permutation}
Shuffling different time slices of data in order to perform \gls{da} is a method that generates new data patterns. It was proposed in \cite{um2017data}, where a fixed slice window was defined from which the data is rearranged, but it has also been applied with variable windows, as it was done in \cite{pan2020data} using \gls{ecg} data at 300 Hz. The main problem of applying permutation is that it does not preserve time dependencies; thus, it can lead to invalid samples. The permutation algorithm can be denoted as follows:
\begin{equation}
    x^{\prime}(w)=\left\{x_{i}, \ldots, x_{i+w}, \ldots, x_{j}, \ldots, x_{j+w}, \ldots, x_{k}, \ldots, x_{k+w}\right\}
\end{equation}
where $ i,j,k $ represents the first index slice of each window, so that each is selected exactly once, and $ w $ denotes the window size if the slices are uniform $ w = T/n $ where $ n $ is the number of total slices.

\subsubsection{Channel permutation}
Changing the position of different channels in multi-dimensional data is a common practice. In computer vision, it is quite popular to swap the RGB channels to perform \gls{da}\cite{lee2019rethinking}. With respect to time series, channel permutation can be applied as long as each channel of the data is still valid. The channel permutation algorithm, for multidimensional data such as $ x=\left\{\left\{x_{11}, \cdots, x_{1T}\right\}, \cdots, \left\{x_{c1}, \cdots, x_{cT} \right\}\right\} $  where $ c $ is the number of channels, is given by
\begin{equation}
    x=\left\{\left\{x_{\sigma(1)1}, \ldots, x_{\sigma(1)T}\right\}, \cdots, \left\{x_{\sigma(c)1}, \ldots, x_{\sigma(c)T} \right\}\right\},
\end{equation}
where $\sigma: \{1, \ldots, c\} \to \{1, \ldots, c\}$ is the used permutation of the channels.

In the time series domain, this algorithm is not applicable to the application of the data, because permutation assumes that the channel information is independent of the channel itself. In other words, the information about the channels is not linked to the particular channel.

That is, in \cite{fu2020data}, they applied this algorithm by flipping the position of the sensors that recorded the data signals, recording the data at 20 Hz using a window of 6 seconds. In the article, the researchers used an exercise mat with eight proximity sensors that they flipped to generate new data. That is, in practice, changing the position of the signal channels.

\subsubsection{Summary of the traditional algorithms}
Figure \ref{figure:TraditionalAlgorithms} shows an example of each algorithm reviewed:

\begin{figure}[H]
    \centering
    \begin{subfigure}{.24\textwidth}
        \centering
        \includegraphics[width=\textwidth]{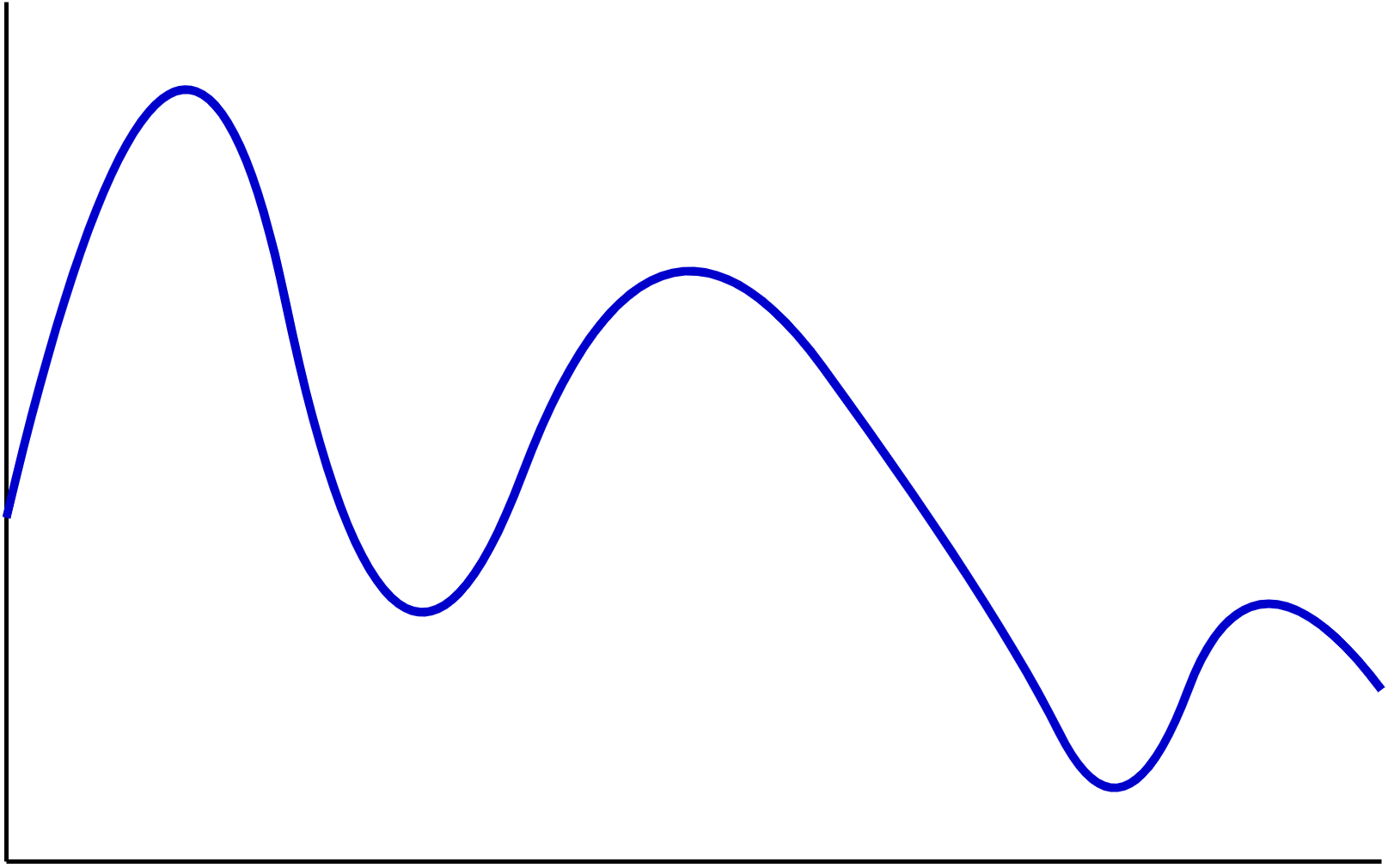}  
        \caption{Ground truth}
        \label{figure:GroundTruth}
    \end{subfigure}
    \begin{subfigure}{.26\textwidth}
        \centering
        \includegraphics[width=\textwidth]{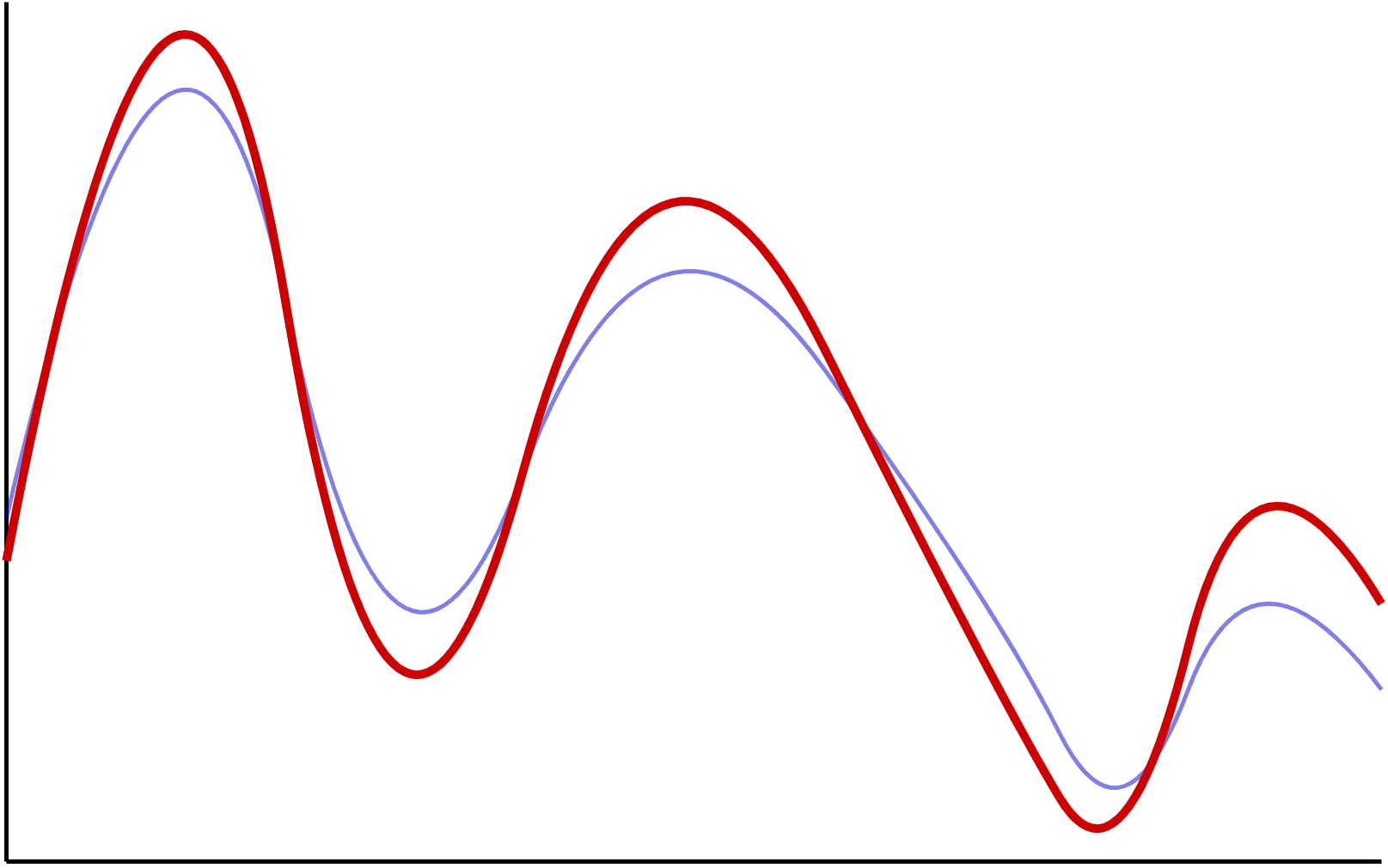}
        \caption{Homogeneous scaling}
        \label{figure:HomogeneousScaling}
    \end{subfigure}
    \begin{subfigure}{.24\textwidth}
        \centering
        \includegraphics[width=\textwidth]{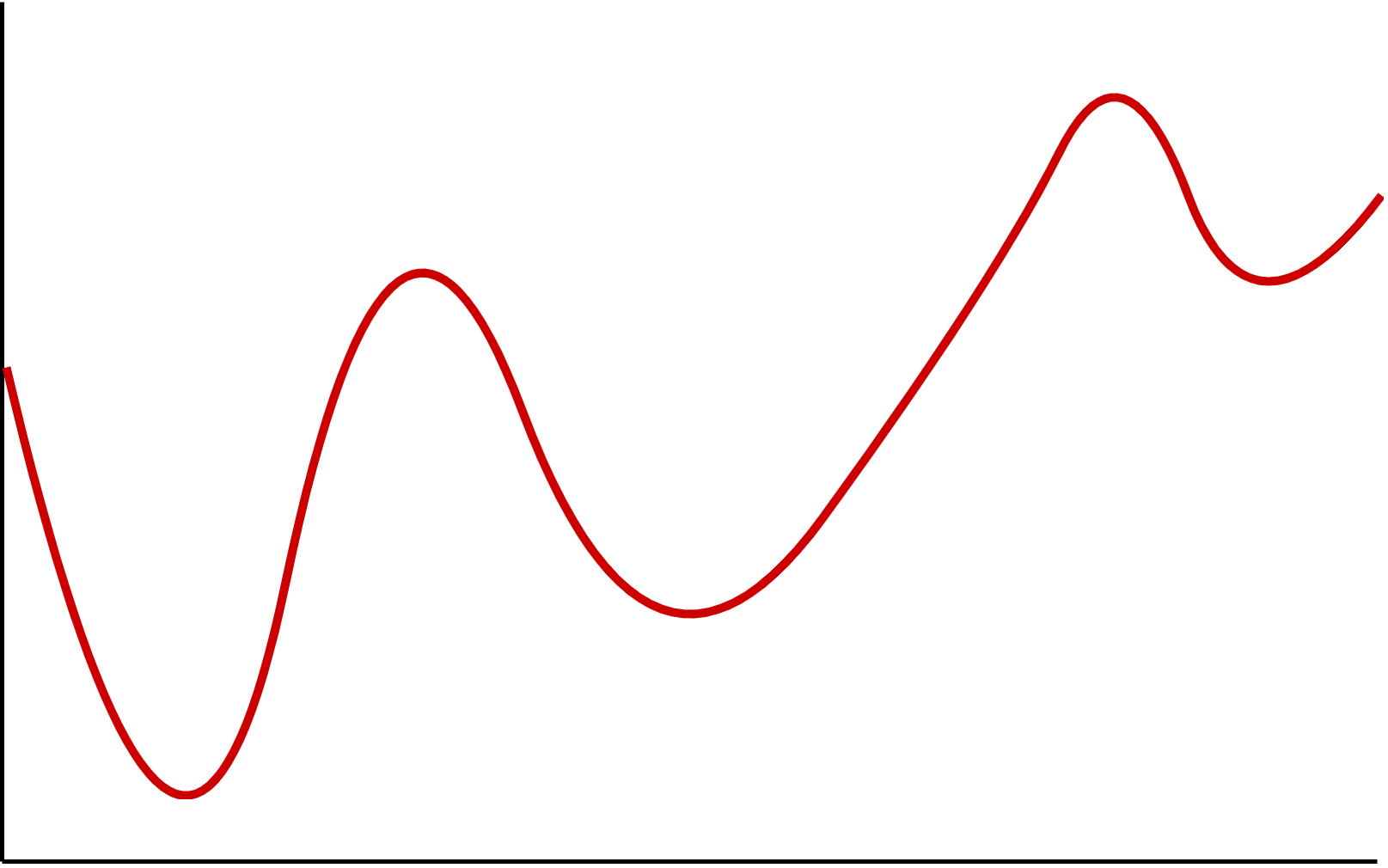}
        \caption{Rotation}
        \label{figure:Rotation}
    \end{subfigure}

\bigskip

    \begin{subfigure}{.24\textwidth}
        \centering
        \includegraphics[width=\textwidth]{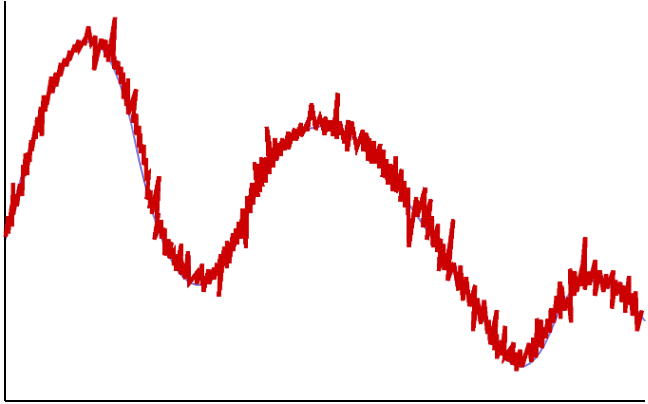}
        \caption{Jittering}
        \label{figure:Jittering}
    \end{subfigure}
    \begin{subfigure}{.26\textwidth}
        \centering
        \includegraphics[width=\textwidth]{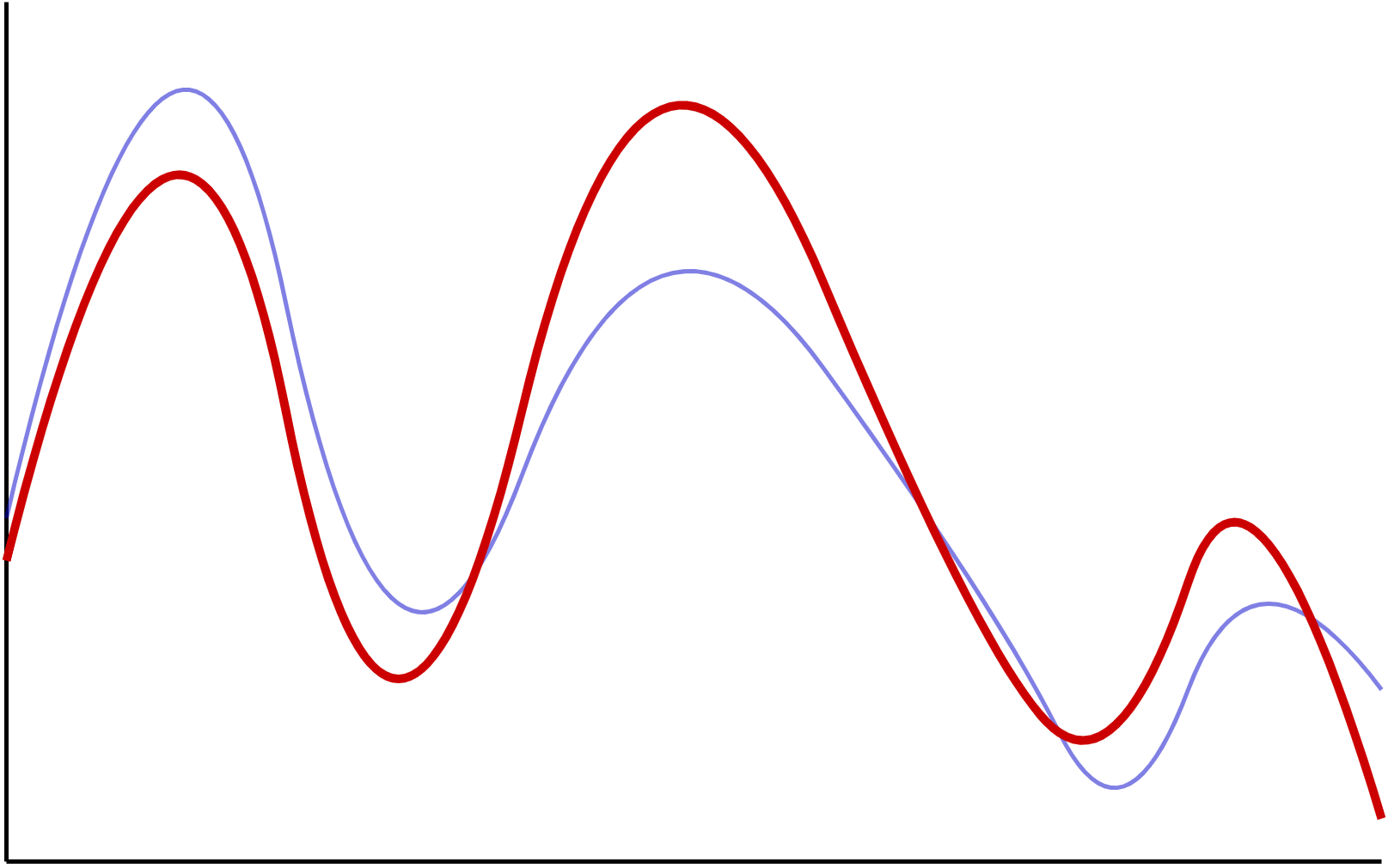}
        \caption{Magnitude warping}
        \label{figure:MagnitudeWarping}
    \end{subfigure}
    \begin{subfigure}{.24\textwidth}
        \centering
        \includegraphics[width=\textwidth]{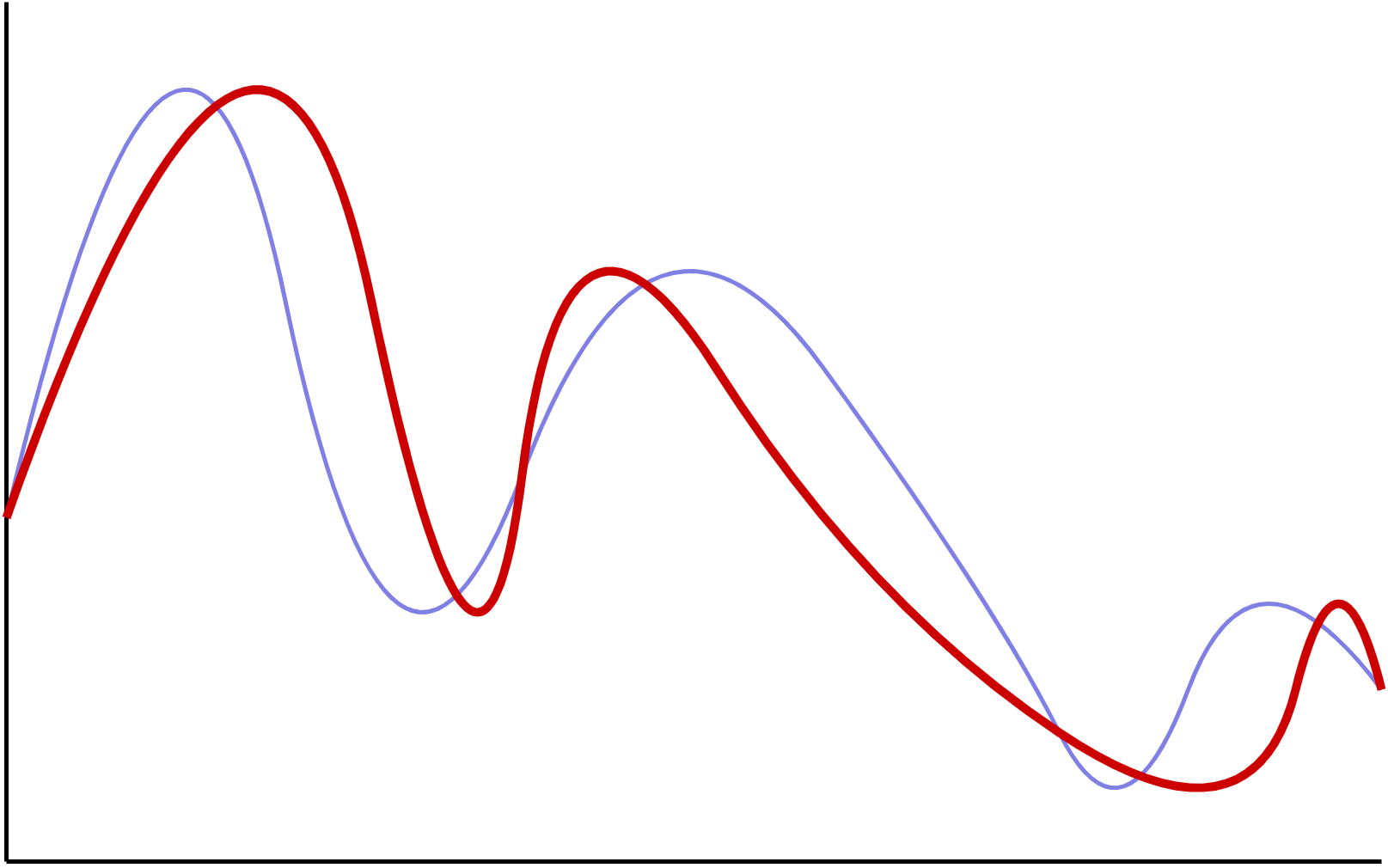}
        \caption{Time warping}
        \label{figure:TimeWarping}
    \end{subfigure}
\end{figure}

\begin{figure}[H] \ContinuedFloat
    \centering
    \begin{subfigure}{.24\textwidth}
        \centering
        \includegraphics[width=\textwidth]{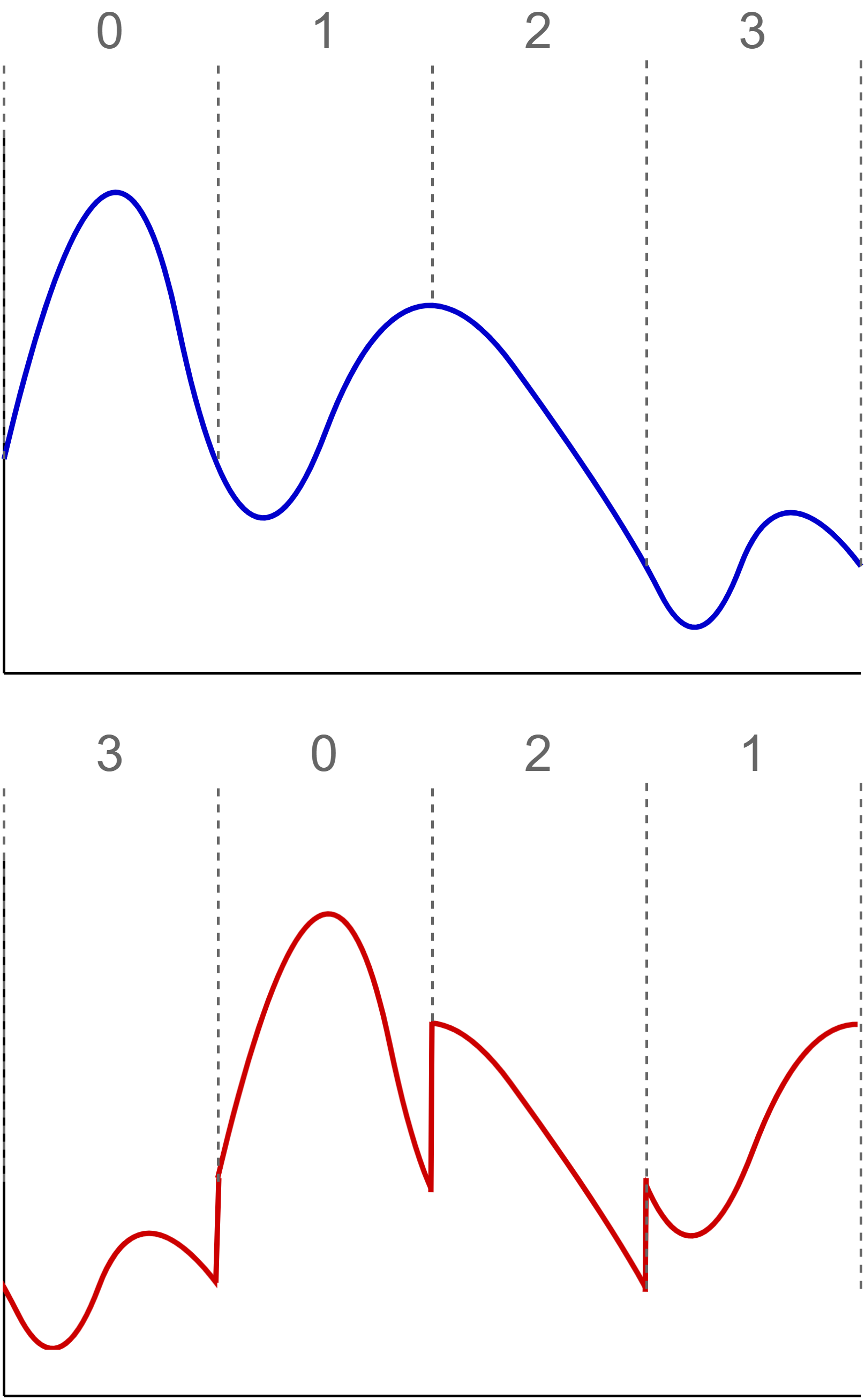}
        \caption{Permutation}
        \label{figure:Permutation}
    \end{subfigure}
    \begin{subfigure}{.26\textwidth}
        \centering
        \includegraphics[width=\textwidth]{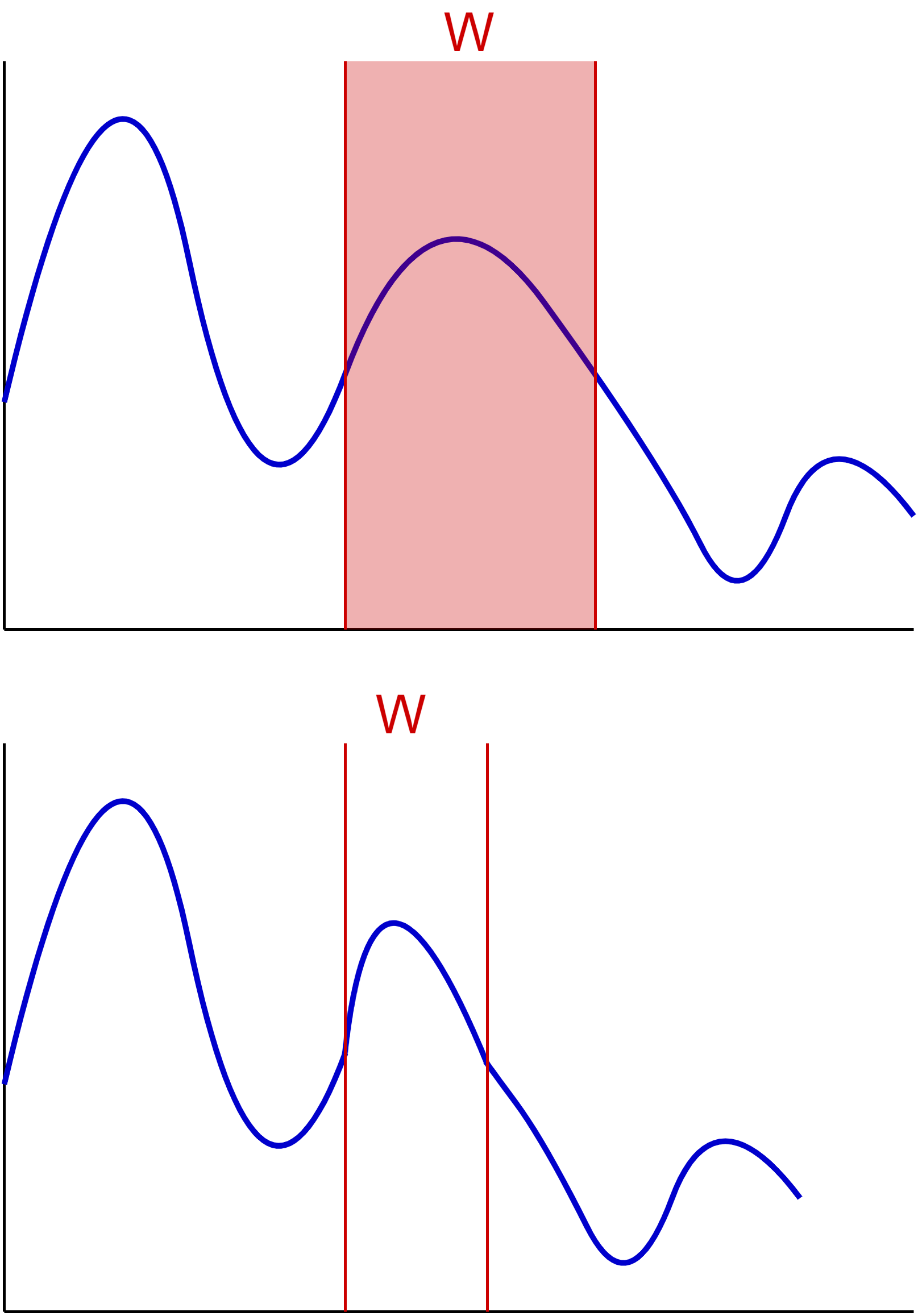}
        \caption{Window warping}
        \label{figure:WindowWarping}
    \end{subfigure}
    \begin{subfigure}{.26\textwidth}
        \centering
        \includegraphics[width=\textwidth]{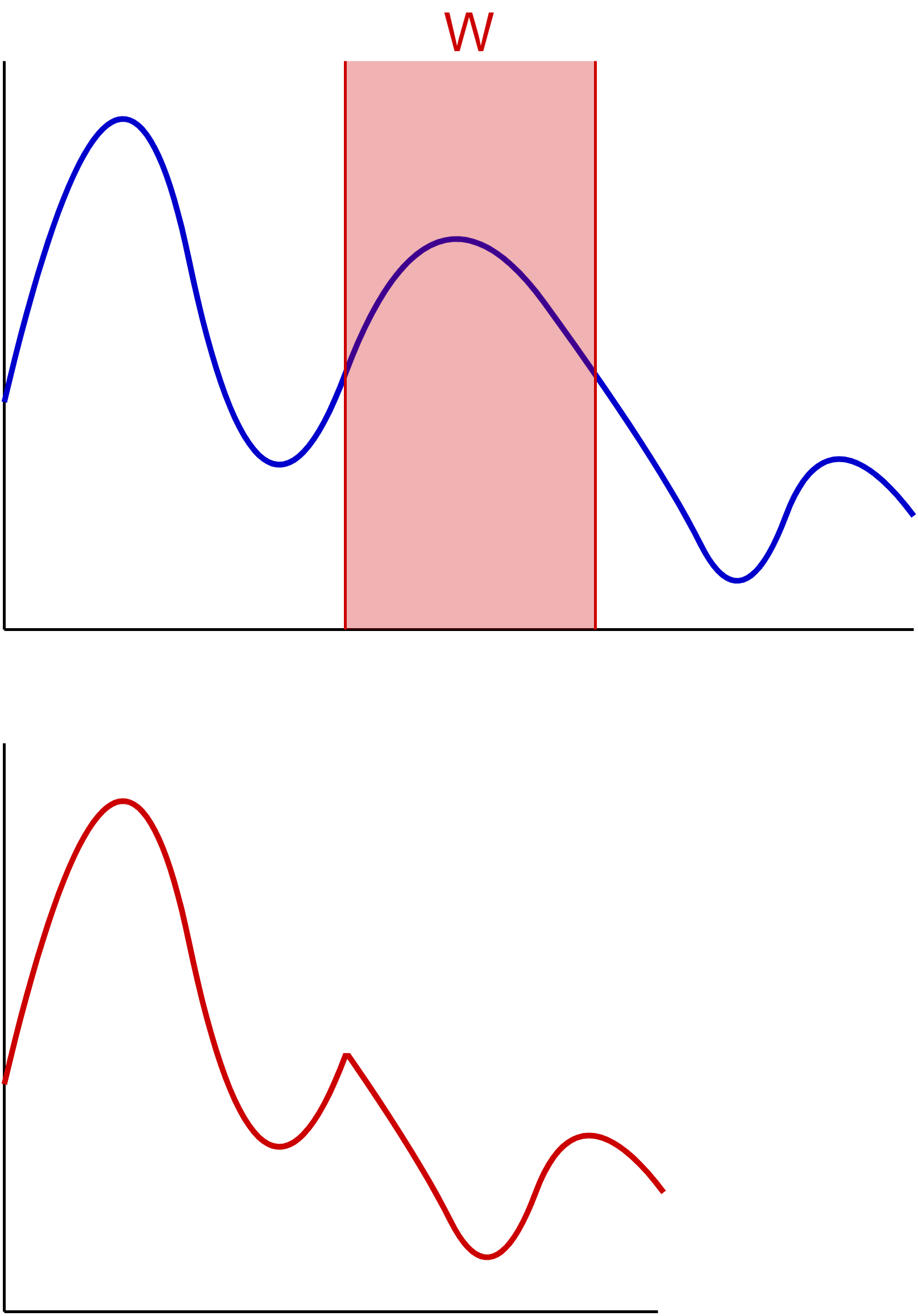}
        \caption{Time Slicing Window}
        \label{figure:TimeSlicingWindow}
    \end{subfigure}
\end{figure}
    
\begin{figure}[H] \ContinuedFloat
    \centering
    \begin{subfigure}{.4\textwidth}
        \centering
        \includegraphics[width=\textwidth]{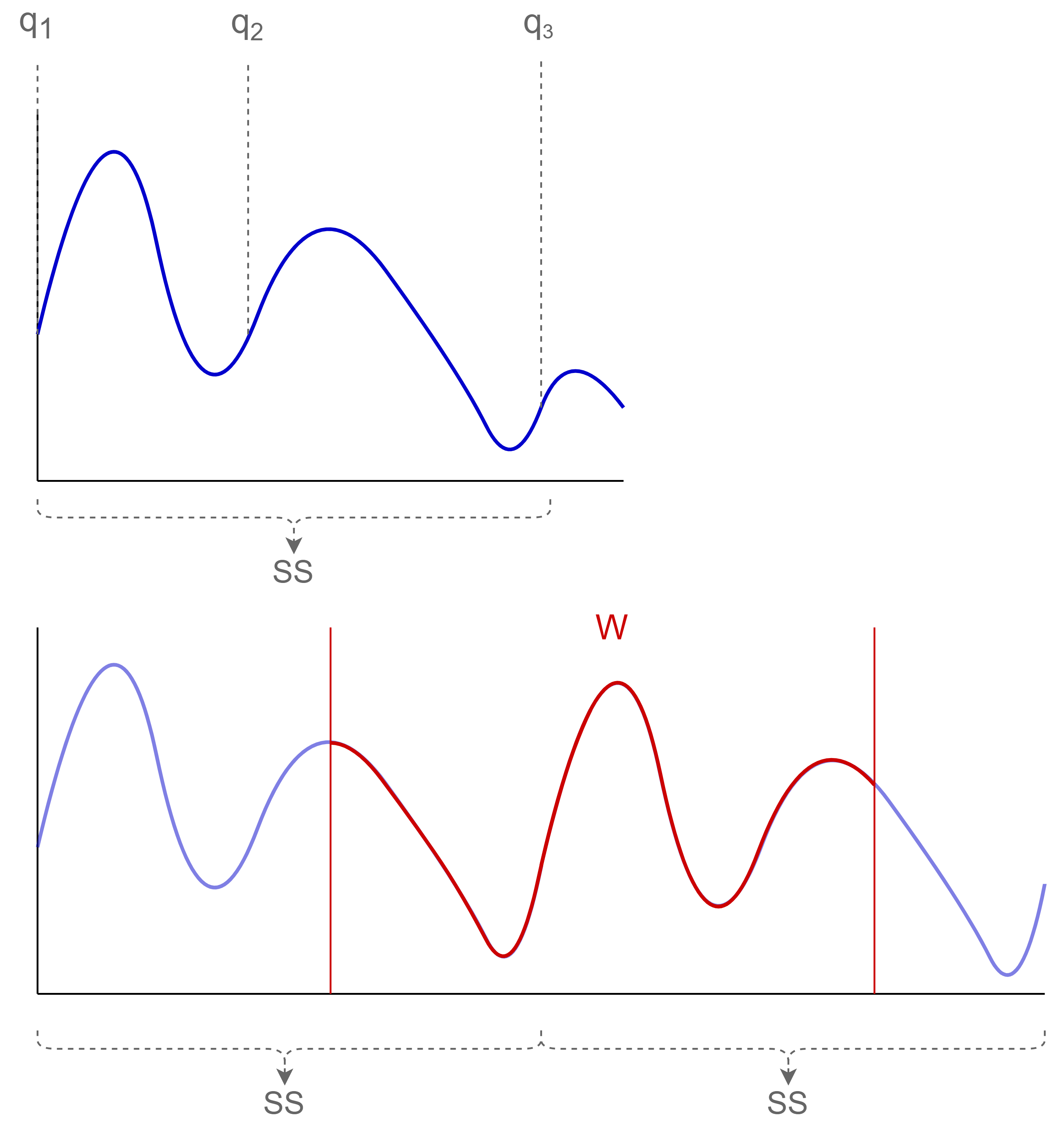}
        \caption{Concatenating resampling}
        \label{figure:ConcatenatingResampling}
    \end{subfigure}
    \begin{subfigure}{.3\textwidth}
        \centering
        \includegraphics[width=\textwidth]{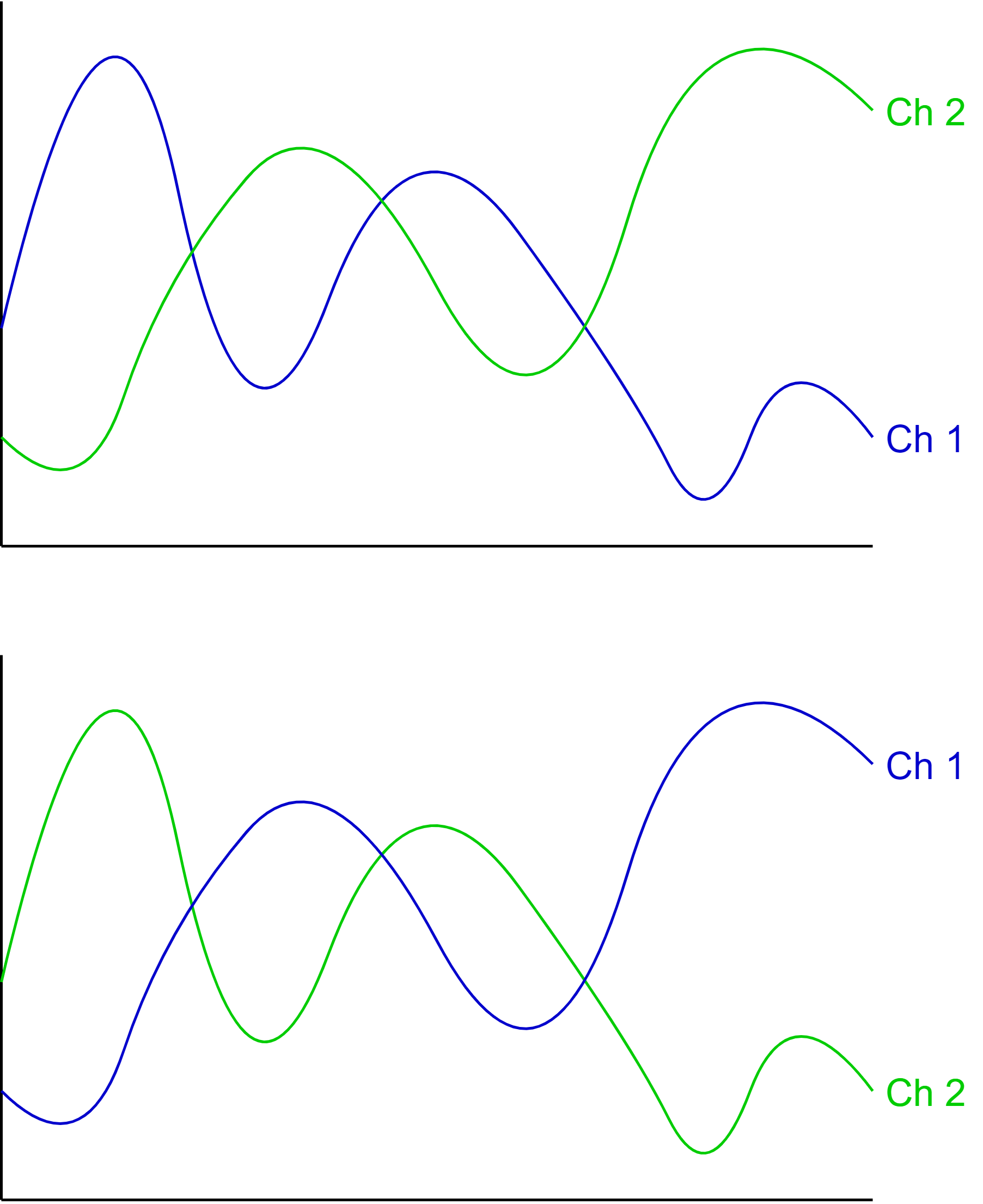}
        \caption{Channel permutation}
        \label{figure:ChannelPermutation}
    \end{subfigure}
    
\caption{Summary of traditional algorithms.}
\label{figure:TraditionalAlgorithms}
\end{figure}

\subsection{Data Augmentation through \gls{vae}} \label{section:vae}
The use of \gls{ae} architectures is nothing more than the evolution of data generation algorithms to produce more and better data, which means that, better, they are varied and therefore the standard deviation with respect to the original data is perfect. To precisely control the deviation of the data, \gls{vae} arises as the evolution of \gls{ae} to generate better synthetic data, as shown in \cite{VAE1} where \gls{vae} is used to generate data for anomaly detection problems with \gls{lstm}. Or this other work \cite{VAE2}, in which they use a dataset augmented with \gls{vae} to improve the recognition of human activity with \gls{lstm}. Even more exhaustive studies \cite{VAE3, VAE4}, show the efficiency of these algorithms in increasing the size of datasets.

But the use of VAEs for \gls{da} is not only intended for neural network models, but can also improve results when traditional Machine Learning algorithms are applied \cite{VAEML}. However, they can also be used in applications with unsupervised training, that is, in \cite{VAEUnsupervised}, which applies them to unsupervised domain adaptation for robust speech recognition.

In \cite{AE_VAE_WGANforTS} they point out that most data augmentation methods for time series use feature space transformations to artificially enlarge the training set; they propose a composition of autoencoders (\gls{ae}s), variational autoencoders (\gls{vae}s) and Wasserstein generative adversarial networks with gradient penalty (WGAN-GPs) for time series augmentation.

In the end, each \gls{vae} model and its hyperparameter configuration make them specialise in the area or format of the dataset they want to work on; but above all to the type of problem for which it will be used afterwards. That is to say, what makes the difference between the models is what the generated data will be used for, regarding problems such as: classification, forecasting, value imputation or prediction.

\subsubsection{Taxonomy for the \gls{vae} algorithms reviewed}

Figure \ref{figure:VAE_Taxonomy} shows a scheme to group the different investigations reviewed in Section \ref{section:vae}, this way all the \gls{vae} algorithms for \gls{da} can be viewed schematically.

\tikzset{every picture/.style={line width=0.75pt}} 
\begin{figure}
\begin{tikzpicture}[x=0.75pt,y=0.75pt,yscale=-1,xscale=1]

\draw    (167,158.16) -- (227.02,51.49) ;
\draw [shift={(228,49.75)}, rotate = 119.37] [color={rgb, 255:red, 0; green, 0; blue, 0 }  ][line width=0.75]    (10.93,-3.29) .. controls (6.95,-1.4) and (3.31,-0.3) .. (0,0) .. controls (3.31,0.3) and (6.95,1.4) .. (10.93,3.29)   ;
\draw    (167,158.16) -- (226.02,104.59) ;
\draw [shift={(227.5,103.25)}, rotate = 137.77] [color={rgb, 255:red, 0; green, 0; blue, 0 }  ][line width=0.75]    (10.93,-3.29) .. controls (6.95,-1.4) and (3.31,-0.3) .. (0,0) .. controls (3.31,0.3) and (6.95,1.4) .. (10.93,3.29)   ;
\draw    (167,158.16) -- (225.57,215.26) ;
\draw [shift={(227,216.66)}, rotate = 224.27] [color={rgb, 255:red, 0; green, 0; blue, 0 }  ][line width=0.75]    (10.93,-3.29) .. controls (6.95,-1.4) and (3.31,-0.3) .. (0,0) .. controls (3.31,0.3) and (6.95,1.4) .. (10.93,3.29)   ;
\draw    (167,158.16) -- (226.5,159.12) ;
\draw [shift={(228.5,159.16)}, rotate = 180.93] [color={rgb, 255:red, 0; green, 0; blue, 0 }  ][line width=0.75]    (10.93,-3.29) .. controls (6.95,-1.4) and (3.31,-0.3) .. (0,0) .. controls (3.31,0.3) and (6.95,1.4) .. (10.93,3.29)   ;
\draw    (167,158.16) -- (227.04,268.4) ;
\draw [shift={(228,270.16)}, rotate = 241.43] [color={rgb, 255:red, 0; green, 0; blue, 0 }  ][line width=0.75]    (10.93,-3.29) .. controls (6.95,-1.4) and (3.31,-0.3) .. (0,0) .. controls (3.31,0.3) and (6.95,1.4) .. (10.93,3.29)   ;
\draw    (201.33,340.74) -- (226.68,323.53) ;
\draw [shift={(228.33,322.41)}, rotate = 145.82] [color={rgb, 255:red, 0; green, 0; blue, 0 }  ][line width=0.75]    (10.93,-3.29) .. controls (6.95,-1.4) and (3.31,-0.3) .. (0,0) .. controls (3.31,0.3) and (6.95,1.4) .. (10.93,3.29)   ;
\draw    (201.33,340.74) -- (226.8,356.6) ;
\draw [shift={(228.5,357.66)}, rotate = 211.91] [color={rgb, 255:red, 0; green, 0; blue, 0 }  ][line width=0.75]    (10.93,-3.29) .. controls (6.95,-1.4) and (3.31,-0.3) .. (0,0) .. controls (3.31,0.3) and (6.95,1.4) .. (10.93,3.29)   ;
\draw    (182,410.41) -- (226.97,438.68) ;
\draw [shift={(228.67,439.74)}, rotate = 212.15] [color={rgb, 255:red, 0; green, 0; blue, 0 }  ][line width=0.75]    (10.93,-3.29) .. controls (6.95,-1.4) and (3.31,-0.3) .. (0,0) .. controls (3.31,0.3) and (6.95,1.4) .. (10.93,3.29)   ;
\draw    (182,410.41) -- (226.8,393.45) ;
\draw [shift={(228.67,392.74)}, rotate = 159.26] [color={rgb, 255:red, 0; green, 0; blue, 0 }  ][line width=0.75]    (10.93,-3.29) .. controls (6.95,-1.4) and (3.31,-0.3) .. (0,0) .. controls (3.31,0.3) and (6.95,1.4) .. (10.93,3.29)   ;

\draw    (17,147.3) -- (165,147.3) -- (165,172.3) -- (17,172.3) -- cycle  ;
\draw (20,151.3) node [anchor=north west][inner sep=0.75pt]   [align=left] {\gls{vae} for classification};
\draw  [color={rgb, 255:red, 74; green, 144; blue, 226 }  ,draw opacity=1 ]  (229.13,26) -- (481.13,26) -- (481.13,72) -- (229.13,72) -- cycle  ;
\draw (232.13,30) node [anchor=north west][inner sep=0.75pt]   [align=left] {Enhancing human activity regonition\\using time series augmented data \cite{VAE2}};
\draw  [color={rgb, 255:red, 74; green, 144; blue, 226 }  ,draw opacity=1 ]  (229.13,82.76) -- (481.13,82.76) -- (481.13,128.76) -- (229.13,128.76) -- cycle  ;
\draw (232.13,86.76) node [anchor=north west][inner sep=0.75pt]   [align=left] {Intelligent random noise modeling by\\improved \gls{vae} \cite{VAE3}};
\draw  [color={rgb, 255:red, 74; green, 144; blue, 226 }  ,draw opacity=1 ]  (229.13,139.52) -- (481.13,139.52) -- (481.13,185.52) -- (229.13,185.52) -- cycle  ;
\draw (232.13,143.52) node [anchor=north west][inner sep=0.75pt]   [align=left] {Improving classification accuracy using\\data DA on small datasets \cite{VAE4}};
\draw  [color={rgb, 255:red, 74; green, 144; blue, 226 }  ,draw opacity=1 ]  (229.13,196.28) -- (481.13,196.28) -- (481.13,242.28) -- (229.13,242.28) -- cycle  ;
\draw (232.13,200.28) node [anchor=north west][inner sep=0.75pt]   [align=left] {Using \gls{vae} to augment sparse time\\series datasets \cite{VAEML}};
\draw  [color={rgb, 255:red, 74; green, 144; blue, 226 }  ,draw opacity=1 ]  (229.13,253.04) -- (481.13,253.04) -- (481.13,299.04) -- (229.13,299.04) -- cycle  ;
\draw (232.13,257.04) node [anchor=north west][inner sep=0.75pt]   [align=left] {Unsupervised domain adaptation for\\robust speech regonition via \gls{vae} \cite{VAEUnsupervised}};
\draw  [color={rgb, 255:red, 74; green, 144; blue, 226 }  ,draw opacity=1 ]  (229.13,309.8) -- (481.13,309.8) -- (481.13,334.8) -- (229.13,334.8) -- cycle  ;
\draw (232.13,313.8) node [anchor=north west][inner sep=0.75pt]   [align=left] {\acrshort{cvae} \acrshort{ecg} anomaly detection \cite{VAE1}};
\draw  [color={rgb, 255:red, 74; green, 144; blue, 226 }  ,draw opacity=1 ]  (229.13,345.56) -- (481.13,345.56) -- (481.13,370.56) -- (229.13,370.56) -- cycle  ;
\draw (232.13,349.56) node [anchor=north west][inner sep=0.75pt]   [align=left] {\acrshort{sisvae} \cite{li2020anomaly}};
\draw  [color={rgb, 255:red, 74; green, 144; blue, 226 }  ,draw opacity=1 ]  (229.13,381.32) -- (481.13,381.32) -- (481.13,406.32) -- (229.13,406.32) -- cycle  ;
\draw (232.13,385.32) node [anchor=north west][inner sep=0.75pt]   [align=left] {GlowImp \cite{liu2021glowimp}};
\draw  [color={rgb, 255:red, 74; green, 144; blue, 226 }  ,draw opacity=1 ]  (229.13,417.09) -- (481.13,417.09) -- (481.13,463.09) -- (229.13,463.09) -- cycle  ;
\draw (232.13,421.09) node [anchor=north west][inner sep=0.75pt]   [align=left] {\gls{vae} based on the shift correction for\\imputation \cite{li2021variational}};
\draw    (17,328.3) -- (201,328.3) -- (201,353.3) -- (17,353.3) -- cycle  ;
\draw (20,332.3) node [anchor=north west][inner sep=0.75pt]   [align=left] {\gls{vae} for anomaly detection};
\draw    (17,397.3) -- (182,397.3) -- (182,422.3) -- (17,422.3) -- cycle  ;
\draw (20,401.3) node [anchor=north west][inner sep=0.75pt]   [align=left] {\gls{vae} for data imputation};
\end{tikzpicture}

\caption{Taxonomy of the presented \gls{vae} architectures.}
\label{figure:VAE_Taxonomy}
\end{figure}
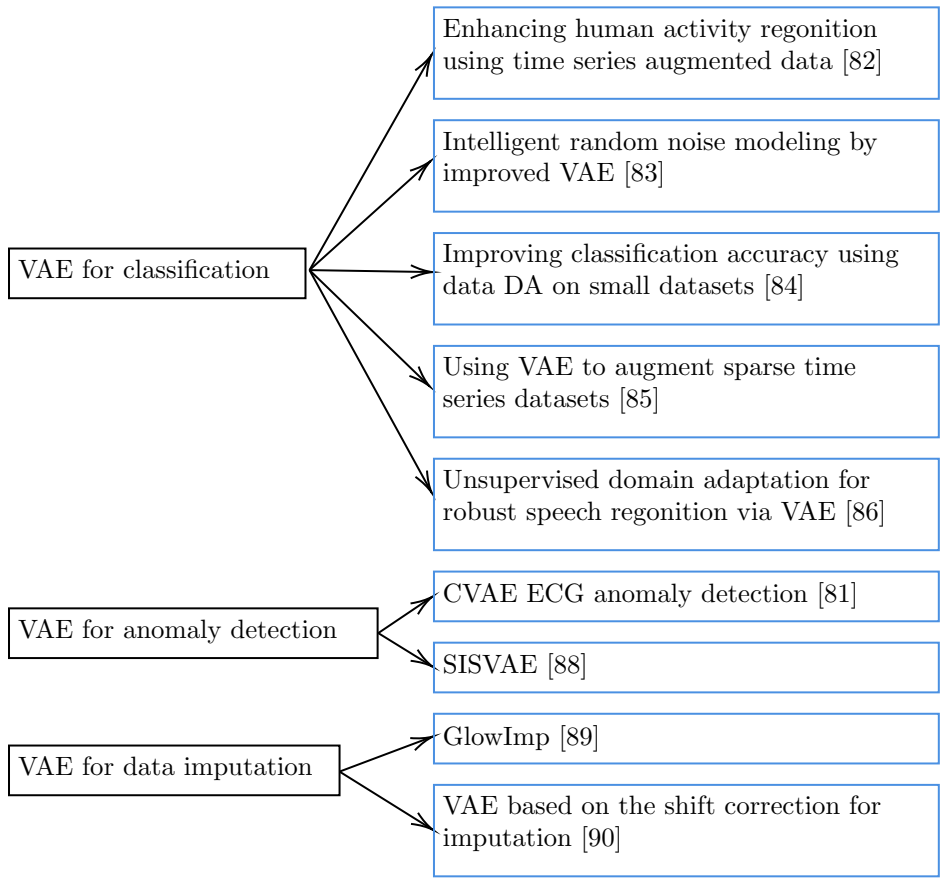

\subsubsection{\glspl{vae} for anomaly detection}
As mentioned before, \glspl{vae} are a \gls{da} architecture that has been widely used in the field of anomaly detection. The main objective of using these models in anomaly detection tasks is to be able to generate data in order to avoid the lack of invalid data of the datasets. The most common scenario is that there are not enough available anomalous samples in order to train machine learning models with the dataset, so the use of \gls{vae} is focused on generating new data.

The work presented in \cite{VAE1} is centred on the classification of \gls{ecg} signals, distinguishing between the ones with cardiac dysfunction. The data used consists of windows of 3600 samples downsampled to 905 sample points during a period of time of 10 seconds. In order to augment the data available it is used a \gls{cvae}\cite{sohn2015learning} that is able to learn which samples are normal and which are anomalous. This \gls{cvae} architecture is composed of \gls{lstm} layers\cite{hochreiter1997long} which process the temporal data of the \gls{ecg} signals.

Another architecture based on the anomaly detection problem is the \gls{sisvae}\cite{li2020anomaly} which uses a \gls{vae} with recurrent layers to maintain temporal dependencies. This work focuses on the problem of abrupt changes between time steps, which led to non-smooth reconstructions of the input data of the model, and therefore temporal abrupt changes in the synthesised samples. The mechanism to avoid this is to introduce a corrective bias for each time step of the signal, calculated using the \gls{kldivergence} \cite{kullback1951information} between one point and the next one in the series. The results of the work are tested using two different time series synthetic datasets.

\subsubsection{\glspl{vae} for data imputation}
One field where the \gls{vae} architecture has been widely used is in data imputation tasks. This process consists of generating new data in a sample where there is missing information. In the temporal series domain, this process is usually used to fill gaps in temporal spaces where there are no available data. In this sense, \glspl{vae} generates synthetic information on demand to fill these gaps, generating new information following the distribution of the original data.

\begin{figure}[h]
    \centering
    \includegraphics[width=.7\textwidth]{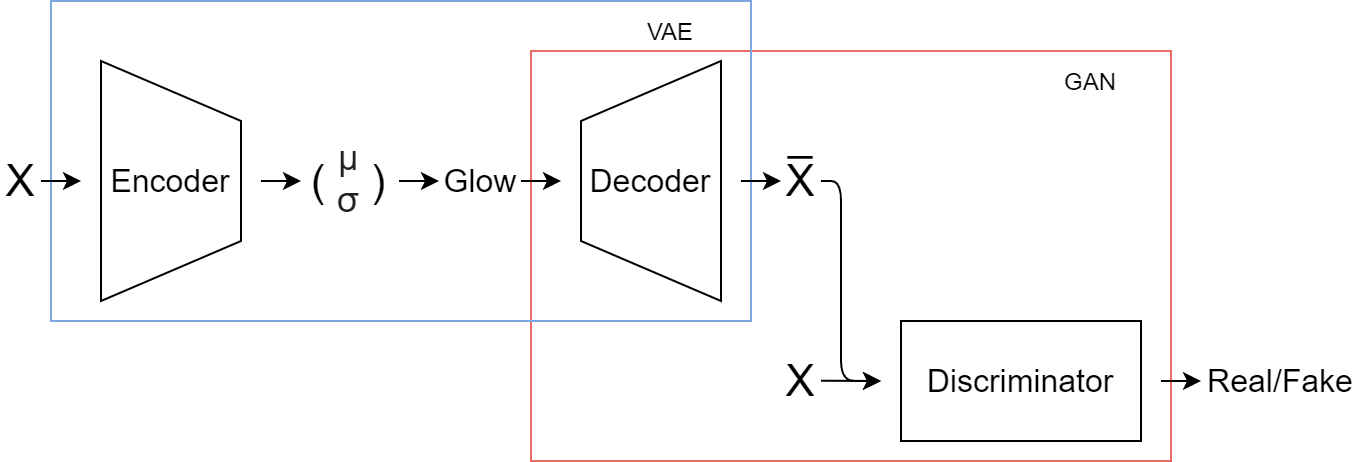}
    \caption{GlowImp architecture. Based on the figure of \cite{liu2021glowimp}.}
    \label{figure:GlowImp_Architecture}
\end{figure}

The GlowImp architecture\cite{liu2021glowimp} was proposed as a combination of the \gls{wgan}\cite{arjovsky2017wasserstein} architecture together with a \gls{vae} to impute missing data. The architecture is composed of the so-called Glow \gls{vae}, which incorporates a function that takes the latent distribution of the traditional \gls{vae} encoder and interpolates the missing values via the Glow Model. The other main part of the architecture is the \gls{gan} model where the generator corresponds with the decoder of the \gls{vae} and the discriminator forces the system to produce realistic samples. The results of the model are tested using two different datasets, the KDD Cup Challenge 2018 dataset containing air quality weather data and the PhysioNet Challenge 2012 which is a collection of multivariate clinical time series data. The architecture of the GlowImp can be seen in figure \ref{figure:GlowImp_Architecture}.

The work of Li et al.\cite{li2021variational} presents a \gls{vae} architecture to impute temporal values using meteorologic datasets. In order to fill in the missing values of the data samples, shift correction is used. This correction tries to counteract the deviation caused by the missing values. This correction is used in the Gaussian latent distribution, where a shift hyperparameter $ \lambda $ is applied which is manually set to center the latent distribution, thus correcting the possible bias produced by missing values. The \gls{vae} architecture used in this work to impute the missing values is $\beta$-\gls{vae}\cite{higgins2016beta}.

\subsection{Data Augmentation through \gls{gan}}
\glspl{gan} are one of the most popular generative models of the last decade, since its introduction in 2014 by Ian Goodfellow\cite{goodfellow2014} this generative architecture has positioned itself as one of the main algorithms for \gls{da}. The main strength of the \gls{gan} architecture is that it learns the distribution of the data by extracting the main features of the samples, without copying the distribution directly. This is known as data generation and its main strength with respect to data augmentation is that it synthesises completely new samples, in contrast with other techniques where the original samples where transformed to generate new instances. This fosters the generalisation and creativity of the synthetic data generated by the model. It is also an important factor that the training of the networks is unsupervised, not necessarily to have labelled data to learn the distribution.

\subsubsection{Taxonomy for the \gls{gan} algorithms reviewed}
Figure \ref{figure:GAN_Taxonomy} shows a scheme to group the different research reviews in Section \ref{section:GAN}, this way all the \gls{gan} algorithms for \gls{da} can be viewed schematically.

\tikzset{every picture/.style={line width=0.75pt}} 
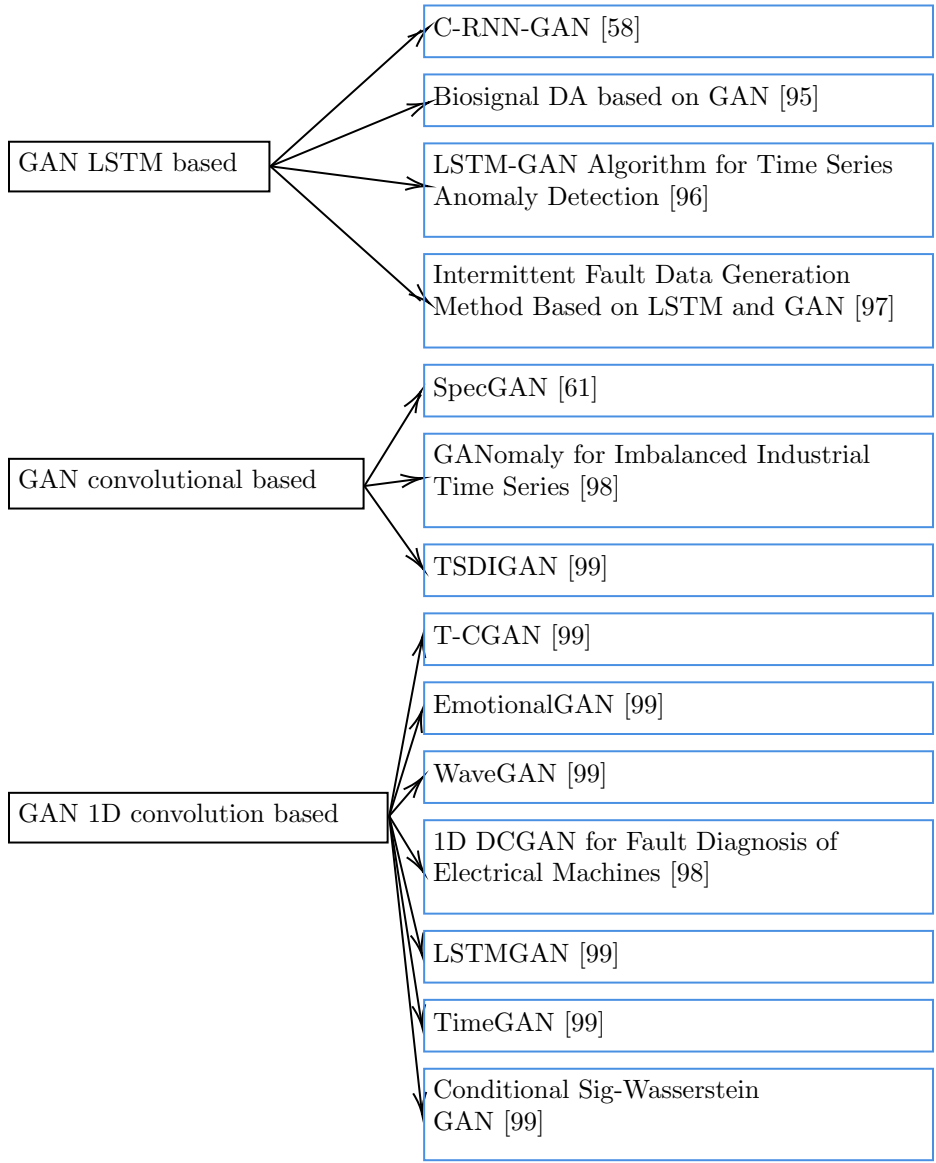
\begin{figure}
\begin{tikzpicture}[x=0.75pt,y=0.75pt,yscale=-1,xscale=1]

\draw    (150.5,133.16) -- (225.9,64.63) ;
\draw [shift={(229.4,63.31)}, rotate = 138.48] [color={rgb, 255:red, 0; green, 0; blue, 0 }  ][line width=0.75]    (10.93,-3.29) .. controls (6.95,-1.4) and (3.31,-0.3) .. (0,0) .. controls (3.31,0.3) and (6.95,1.4) .. (10.93,3.29)   ;
\draw    (150.5,133.16) -- (225.95,101.66) ;
\draw [shift={(229.8,100.91)}, rotate = 157.87] [color={rgb, 255:red, 0; green, 0; blue, 0 }  ][line width=0.75]    (10.93,-3.29) .. controls (6.95,-1.4) and (3.31,-0.3) .. (0,0) .. controls (3.31,0.3) and (6.95,1.4) .. (10.93,3.29)   ;
\draw    (150.5,133.16) -- (225.49,200.85) ;
\draw [shift={(230,202.16)}, rotate = 220.96] [color={rgb, 255:red, 0; green, 0; blue, 0 }  ][line width=0.75]    (10.93,-3.29) .. controls (6.95,-1.4) and (3.31,-0.3) .. (0,0) .. controls (3.31,0.3) and (6.95,1.4) .. (10.93,3.29)   ;
\draw    (150.5,133.16) -- (225.02,142.9) ;
\draw [shift={(229,143.16)}, rotate = 187.26] [color={rgb, 255:red, 0; green, 0; blue, 0 }  ][line width=0.75]    (10.93,-3.29) .. controls (6.95,-1.4) and (3.31,-0.3) .. (0,0) .. controls (3.31,0.3) and (6.95,1.4) .. (10.93,3.29)   ;
\draw    (197.4,293.98) -- (225.65,247.96) ;
\draw [shift={(225.67,246.24)}, rotate = 120.63] [color={rgb, 255:red, 0; green, 0; blue, 0 }  ][line width=0.75]    (10.93,-3.29) .. controls (6.95,-1.4) and (3.31,-0.3) .. (0,0) .. controls (3.31,0.3) and (6.95,1.4) .. (10.93,3.29)   ;
\draw    (197.4,293.98) -- (225.62,290.25) ;
\draw [shift={(226.6,289.98)}, rotate = 172.2] [color={rgb, 255:red, 0; green, 0; blue, 0 }  ][line width=0.75]    (10.93,-3.29) .. controls (6.95,-1.4) and (3.31,-0.3) .. (0,0) .. controls (3.31,0.3) and (6.95,1.4) .. (10.93,3.29)   ;
\draw    (197.4,293.98) -- (225.85,334.34) ;
\draw [shift={(227,335.98)}, rotate = 234.83] [color={rgb, 255:red, 0; green, 0; blue, 0 }  ][line width=0.75]    (10.93,-3.29) .. controls (6.95,-1.4) and (3.31,-0.3) .. (0,0) .. controls (3.31,0.3) and (6.95,1.4) .. (10.93,3.29)   ;
\draw    (209.8,459.98) -- (226.24,371.14) ;
\draw [shift={(226.6,369.18)}, rotate = 100.48] [color={rgb, 255:red, 0; green, 0; blue, 0 }  ][line width=0.75]    (10.93,-3.29) .. controls (6.95,-1.4) and (3.31,-0.3) .. (0,0) .. controls (3.31,0.3) and (6.95,1.4) .. (10.93,3.29)   ;
\draw    (209.8,459.98) -- (225.61,408.69) ;
\draw [shift={(226.2,406.78)}, rotate = 107.13] [color={rgb, 255:red, 0; green, 0; blue, 0 }  ][line width=0.75]    (10.93,-3.29) .. controls (6.95,-1.4) and (3.31,-0.3) .. (0,0) .. controls (3.31,0.3) and (6.95,1.4) .. (10.93,3.29)   ;
\draw    (209.8,459.98) -- (225.3,441.89) ;
\draw [shift={(226.6,440.38)}, rotate = 130.6] [color={rgb, 255:red, 0; green, 0; blue, 0 }  ][line width=0.75]    (10.93,-3.29) .. controls (6.95,-1.4) and (3.31,-0.3) .. (0,0) .. controls (3.31,0.3) and (6.95,1.4) .. (10.93,3.29)   ;
\draw    (209.8,459.98) -- (225.96,486.66) ;
\draw [shift={(227,488.38)}, rotate = 238.8] [color={rgb, 255:red, 0; green, 0; blue, 0 }  ][line width=0.75]    (10.93,-3.29) .. controls (6.95,-1.4) and (3.31,-0.3) .. (0,0) .. controls (3.31,0.3) and (6.95,1.4) .. (10.93,3.29)   ;
\draw    (209.8,459.98) -- (225.74,528.03) ;
\draw [shift={(226.2,529.98)}, rotate = 256.81] [color={rgb, 255:red, 0; green, 0; blue, 0 }  ][line width=0.75]    (10.93,-3.29) .. controls (6.95,-1.4) and (3.31,-0.3) .. (0,0) .. controls (3.31,0.3) and (6.95,1.4) .. (10.93,3.29)   ;
\draw    (209.8,459.98) -- (225.89,564) ;
\draw [shift={(226.2,565.98)}, rotate = 261.21] [color={rgb, 255:red, 0; green, 0; blue, 0 }  ][line width=0.75]    (10.93,-3.29) .. controls (6.95,-1.4) and (3.31,-0.3) .. (0,0) .. controls (3.31,0.3) and (6.95,1.4) .. (10.93,3.29)   ;
\draw    (209.8,459.98) -- (225.98,607.99) ;
\draw [shift={(226.2,609.98)}, rotate = 263.76] [color={rgb, 255:red, 0; green, 0; blue, 0 }  ][line width=0.75]    (10.93,-3.29) .. controls (6.95,-1.4) and (3.31,-0.3) .. (0,0) .. controls (3.31,0.3) and (6.95,1.4) .. (10.93,3.29)   ;

\draw    (20,120.8) -- (150,120.8) -- (150,145.8) -- (20,145.8) -- cycle  ;
\draw (23,124.8) node [anchor=north west][inner sep=0.75pt]   [align=left] {GAN LSTM based};
\draw  [color={rgb, 255:red, 74; green, 144; blue, 226 }  ,draw opacity=1 ]  (227.23,52.27) -- (481.23,52.27) -- (481.23,78.27) -- (227.23,78.27) -- cycle  ;
\draw (230.23,56.27) node [anchor=north west][inner sep=0.75pt]   [align=left] {C-RNN-GAN \cite{mogren2016c-crnngan}};
\draw  [color={rgb, 255:red, 74; green, 144; blue, 226 }  ,draw opacity=1 ]  (227.23,86.95) -- (481.23,86.95) -- (481.23,112.95) -- (227.23,112.95) -- cycle  ;
\draw (230.23,90.95) node [anchor=north west][inner sep=0.75pt]   [align=left] {Biosignal DA based on GAN \cite{haradal2018biosignal}};
\draw  [color={rgb, 255:red, 74; green, 144; blue, 226 }  ,draw opacity=1 ]  (227.23,121.63) -- (481.23,121.63) -- (481.23,168.63) -- (227.23,168.63) -- cycle  ;
\draw (230.23,125.63) node [anchor=north west][inner sep=0.75pt]   [align=left] {LSTM-GAN Algorithm for Time Series\\Anomaly Detection \cite{zhu2019novel}};
\draw  [color={rgb, 255:red, 74; green, 144; blue, 226 }  ,draw opacity=1 ]  (227.23,177.31) -- (481.23,177.31) -- (481.23,224.31) -- (227.23,224.31) -- cycle  ;
\draw (230.23,181.31) node [anchor=north west][inner sep=0.75pt]   [align=left] {Intermittent Fault Data Generation\\Method Based on LSTM and GAN \cite{shi2021intermittent}};
\draw    (20,280.3) -- (197,280.3) -- (197,305.3) -- (20,305.3) -- cycle  ;
\draw (23,284.3) node [anchor=north west][inner sep=0.75pt]   [align=left] {GAN convolutional based};
\draw  [color={rgb, 255:red, 74; green, 144; blue, 226 }  ,draw opacity=1 ]  (227.23,267.67) -- (481.23,267.67) -- (481.23,314.67) -- (227.23,314.67) -- cycle  ;
\draw (230.23,271.67) node [anchor=north west][inner sep=0.75pt]   [align=left] {GANomaly for Imbalanced Industrial\\Time Series \cite{jiang2019gan}};
\draw  [color={rgb, 255:red, 74; green, 144; blue, 226 }  ,draw opacity=1 ]  (227.23,232.99) -- (481.23,232.99) -- (481.23,258.99) -- (227.23,258.99) -- cycle  ;
\draw (230.23,236.99) node [anchor=north west][inner sep=0.75pt]   [align=left] {SpecGAN \cite{donahue2018adversarial_wavegan}};
\draw  [color={rgb, 255:red, 74; green, 144; blue, 226 }  ,draw opacity=1 ]  (227.23,323.35) -- (481.23,323.35) -- (481.23,349.35) -- (227.23,349.35) -- cycle  ;
\draw (230.23,327.35) node [anchor=north west][inner sep=0.75pt]   [align=left] {TSDIGAN \cite{huang2021deep}};
\draw  [color={rgb, 255:red, 74; green, 144; blue, 226 }  ,draw opacity=1 ]  (227.23,358.03) -- (481.23,358.03) -- (481.23,384.03) -- (227.23,384.03) -- cycle  ;
\draw (230.23,362.03) node [anchor=north west][inner sep=0.75pt]   [align=left] {T-CGAN \cite{huang2021deep}};
\draw  [color={rgb, 255:red, 74; green, 144; blue, 226 }  ,draw opacity=1 ]  (227.23,392.71) -- (481.23,392.71) -- (481.23,418.71) -- (227.23,418.71) -- cycle  ;
\draw (230.23,396.71) node [anchor=north west][inner sep=0.75pt]   [align=left] {EmotionalGAN \cite{huang2021deep}};
\draw  [color={rgb, 255:red, 74; green, 144; blue, 226 }  ,draw opacity=1 ]  (227.23,427.39) -- (481.23,427.39) -- (481.23,453.39) -- (227.23,453.39) -- cycle  ;
\draw (230.23,431.39) node [anchor=north west][inner sep=0.75pt]   [align=left] {WaveGAN \cite{huang2021deep}};
\draw  [color={rgb, 255:red, 74; green, 144; blue, 226 }  ,draw opacity=1 ]  (227.23,462.07) -- (481.23,462.07) -- (481.23,509.07) -- (227.23,509.07) -- cycle  ;
\draw (230.23,466.07) node [anchor=north west][inner sep=0.75pt]   [align=left] {1D DCGAN for Fault Diagnosis of\\Electrical Machines \cite{jiang2019gan}};
\draw  [color={rgb, 255:red, 74; green, 144; blue, 226 }  ,draw opacity=1 ]  (227.23,517.75) -- (481.23,517.75) -- (481.23,543.75) -- (227.23,543.75) -- cycle  ;
\draw (230.23,521.75) node [anchor=north west][inner sep=0.75pt]   [align=left] {LSTMGAN \cite{huang2021deep}};
\draw  [color={rgb, 255:red, 74; green, 144; blue, 226 }  ,draw opacity=1 ]  (227.23,552.43) -- (481.23,552.43) -- (481.23,578.43) -- (227.23,578.43) -- cycle  ;
\draw (230.23,556.43) node [anchor=north west][inner sep=0.75pt]   [align=left] {TimeGAN \cite{huang2021deep}};
\draw    (20,448.3) -- (209,448.3) -- (209,473.3) -- (20,473.3) -- cycle  ;
\draw (23,452.3) node [anchor=north west][inner sep=0.75pt]   [align=left] {GAN 1D convolution based};
\draw  [color={rgb, 255:red, 74; green, 144; blue, 226 }  ,draw opacity=1 ]  (227.23,587.09) -- (481.23,587.09) -- (481.23,633.09) -- (227.23,633.09) -- cycle  ;
\draw (230.23,591.09) node [anchor=north west][inner sep=0.75pt]   [align=left] {Conditional Sig-Wasserstein\\GAN \cite{huang2021deep}};
\end{tikzpicture}

\caption{Taxonomy of the presented \gls{gan} architectures.}
\label{figure:GAN_Taxonomy}
\end{figure}

\glsreset{lstm}
\subsubsection{\gls{lstm} based \glspl{gan}}
One of the approaches to adapt the \gls{gan} architecture to time series is to use recurrent networks as the base of \gls{ann}. These \glspl{gan} substitute the regular fully connected or convolutional layers with recurrent layers, able to have memory that links temporal features of the data. The main strength of this set of architectures is that they are able to process this temporal information that the input data have, similar to the spatial information processing of a convolutional neural network.

\gls{crnngan}\cite{mogren2016c-crnngan} is one of the first \gls{gan} architectures proposed specifically for time series data. In particular, using them to learn and synthesize music tracks was proposed in this work. This \glspl{gan} uses \gls{lstm} blocks\cite{hochreiter1997long} as its main learning structure. The learning algorithm is the same as standard training \gls{gan}, where the network generator concatenates each input with the output of the previous cells and the discriminator is made up of a bidirectional recurrent network\cite{schuster1997bidirectional_blstm}. The internal composition of the discriminator is based on the work of Horchreiter\cite{hochreiter1998vanishing} and Bengio et al.\cite{bengio1994learning} that avoids the gradient vanishing and strengths the temporal dependencies. The results of the work are discussed using a dataset of 3697 musical midi files from 160 different composers of classical music with a tick resolution of 384 per quarter note.

The work presented by Haradal et al.\cite{haradal2018biosignal} also proposes a \gls{gan} architecture based on the implementation of \gls{lstm} cells in both the generator and discriminator networks to adapt to time series data. The discriminator output is generated by applying an average pooling to the outputs generated by each layer, averaging the whole data sample into a unique scalar output which corresponds to the probability of the sample being generated by the generator network. This architecture was used to generate \gls{ecg}\cite{olszewski2001generalized} and \gls{eeg}\cite{andrzejak2001indications} data to improve the classification accuracy of a \gls{ann} classifier.

The \gls{lstm} and \gls{gan} combination has also been used for anomaly detection in the work of Zhu et al.\cite{zhu2019novel} where the \gls{lstm} layers are used in the discriminator to extract temporal information from the data, while the \gls{gan} architecture provides the system with the ability to extract the most important features of the data. Training for detecting anomalies in the data has two phases. The first phase, known as the training phase, is a standard \gls{gan} training in which the discriminator learns how to distinguish between real and synthetic data. In the second phase, the so-called testing phase, the training consists of a feature extraction that generates and embedding of the dataset samples, these features are then reconstructed by the generator and compared with the original data, the task of the discriminator is to distinguish the real and the reconstructed data, which is anomalous. This research tested its results using two different datasets, \gls{ecg} data with a window of 96 data values that is training to detect anomalous cases and a dataset with the statistics of taxi traffic in New York City, with 48 points each data sample.

The work presented by Shi et al.\cite{shi2021intermittent} uses the \gls{gan} architecture to generate sequences of faulty data from two different types. Different models are trained for each type. The generator and discriminator of each \gls{gan} are made of a many-to-many \gls{lstm} model that processes the voltage signal data and the sampling length of each step of the sequence. In this way, the generator output is composed of two vectors, one for the voltage and the other for the length, while the discriminator processes these data and its output is generated by averaging the classification of each step and generating a unique binary output.

\subsubsection{Convolutional \glspl{gan} applied to the time series domain}
In order to apply \glspl{gan} to replicate time series, one of the most popular techniques used is to treat the time data as an image. Different approximations have been used in this field, where the focus is on how to transform the data into an image format, rather than adapting the \gls{gan} architecture to process time series information. One of the main advantages of this technique is that it does not have to deal with the design of \gls{gan}, which is a complex process due to the particularities of the architecture. The adaptation of the original data to an image is different in each case. Different works published during the last years will be reviewed in order to study different approximations to this transformation.

An example of this use is the one proposed with SpecGAN\cite{donahue2018adversarial_wavegan} which tries to operate with sound spectrograms that represent audio samples. This approach uses \gls{dcgan}\cite{radford2015unsupervised_dcgan} as the main algorithm for \gls{da}, but prior to that, it processes the audio signal to generate images for each audio track. The process of transforming audio into image "can be approximately inverted" in the author's own words. First, the Fourier transformation is applied to each audio to generate a matrix of the frequencies of the data. Then the scale of the data is adapted logarithmically and normalised to a normal distribution for a better understanding. Finally, the images are clipped to 3 standard deviations and rescaled within the $ [-1, 1] $ range. In particular, the 16384 points of each sample are converted into a 128x128 pixel image. As mentioned above, this process is reversible, so once the new data are generated using \gls{dcgan} they can be transformed to audio data using the reverse process. One advantage of using this process is that it opens up the possibility of comparing different audio generation algorithms treating the results as images; in the original paper, the results of the SpecGAN are compared with the WaveGAN, which is proposed in the same article.

The work presented by Jiang et al.\cite{jiang2019gan} uses the GANomaly architecture\cite{akcay2018ganomaly} to process different time series data. The GANomaly is used for anomaly detection in industrial tasks; it introduces a feature extraction into the network, which pre-processes the input data of both the generator and the generator. The generator is composed of an encoder-decoder-encoder network, which makes it possible to learn the latent representations generated by the feature extraction part. Regarding the data used for training, rolling bearing data was used to detect anomalies, collected at 12 and 48 kHz in two different datasets. The collected data is converted from the time series domain into images by generating a spectrogram, thus converting the time series data into the image domain. In particular, they used Bearing Data from Case Western Reserve University\footnote{https://engineering.case.edu/bearingdatacenter}.

The \gls{tsdigan}\cite{huang2021deep} is an architecture proposed for missing data reconstruction. In particular, traffic data is used consisting on 104,544 traffic records. In this work, \gls{gan} is in charge of generating synthetic data that fill in the missing data gaps with realistic information. The approach used in the paper to treat time-series traffic data is to transform them into an image format using the proposed method called \gls{gasf}. The \gls{gasf} algorithm is focused on maintaining the time dependency of the traffic data; this algorithm is capable of transforming the data into a matrix by representing each time data point to a polar coordinate system within the range $ [-1, 1] $. Then, each point is encoded by its angular cosine and radius. This generates a matrix with the temporal correlation between each point, which is then fed to the networks. Finally, the data are processed using a convolutional-based \gls{gan} that uses its generator to generate new data and reconstruct the missing values.

\subsubsection{1D convolutional \glspl{gan}}
\label{TCGAN}
Temporal \glspl{cnn} are \glspl{cnn}\cite{lecun1998gradient} where the convolutional operation is calculated in 1D instead of traditional 2D convolution. These networks adapt the geometric information captured by the 2D \glspl{cnn} to a temporal domain, lowering the dimensions of the learnt filters to 1D. These networks have been used in works such as \cite{iwana2021empirical} to classify data from temporal series.

In the recent years, different \gls{gan} architectures have been proposed that use these 1D convolutional layers as a base, replacing the traditional 2D convolutions of \glspl{gan} applied to computer vision tasks. In this approximation, it is very straightforward to adapt traditional \gls{gan} architectures to the time series domain, making it very plausible for use in time series-related tasks.

The \gls{tcgan}\cite{ramponi2018t} is a \gls{gan} architecture based on the idea of transforming the \gls{cgan}\cite{mirza2014conditional} architecture to time series domain by replacing the 2D convolutional layers with the 1D convolutional layers. The performance of the model was validated using one synthetic and three real-world datasets. The synthetic data was constructed using sine waves and sawtooth waves. The other datasets consisted of: astronomical light curves of 1024 points per sample, a power demand dataset of samples of 24 points, and an \gls{ecg} dataset with 96 points for each sample.

Emotional\gls{gan}\cite{chen2019emotionalgan} also applies these 1D convolutional layers to create a \gls{gan} architecture to augment an \gls{ecg} dataset improving the classification of \gls{svm} and Random Forest models when classifying the emotions of each subject. This work used different datasets varying their frequency rate between 256 and 2048 Hz.

The work published by Donahue et al.\cite{donahue2018adversarial_wavegan} presents the WaveGAN architecture, which is based on the application of 1D convolutional layers to sound data. This \gls{gan} uses the \gls{dcgan} architecture, but changes the convolutions to 1D. As suggested, these 1D convolutions should have a wider receptive field respecting the 2D convolutions of image processing; this is based on the particularities of the audio data, in which each cycle of a musical note sampled at 16 kHz may take 36 samples to complete. Therefore, it is necessary to use wider filters to capture the distanced temporal dependencies of the data. This feature of the sound data was previously taken into account with solutions such as the dilated convolutions proposed in WaveNet\cite{van2016wavenet}. This enlargement of the receptive field is compensated for by reducing one dimension, changing from 5x5 convolutions to 25 1-dimensional convolutions, and maintaining the number of parameters of the network. The rest of the architecture maintains the standard \gls{gan} architecture, allowing the synthesis of audio tracks with unsupervised training \gls{gan}.

This approximation has also been followed by Sabir et al.\cite{sabir2021signal} for augmenting DC current signals, using samples of current signal with a frequency of 100 Hz during 16 seconds. The proposed work used the \gls{dcgan} architecture as a base and changes the original convolutions to 1D convolutions. In particular, this work has 2 different \glspl{gan}, one that generates healthy signals and the other is in charge of generating faulty data.

There are also hybrid implementations that combine 1D convolutions with other techniques, such as in \cite{zhu2019novel} where \gls{lstm}-\gls{gan} is proposed. This architecture combines the \gls{lstm} cell in the discriminator network with the 1D convolutional layers used in the generator network.

\subsubsection{Time series Generative Adversarial Networks (TimeGAN)}
The \gls{timegan} architecture\cite{yoon2019time_timegan} tries to implement a \gls{gan} model to perform \gls{da} on time-series data, but differentiates itself from other previous alternatives by adding a new loss function that tries to capture the stepwise dependencies of the data. Previous implementations of \glspl{gan} in data sequences were based on the use of recurrent networks for the generator and discriminator networks of \gls{gan}\cite{mogren2016c-crnngan, esteban2017real_rcgan}, but this approximation may not be sufficient to accurately replicate the temporal transitions of the original data.

This work divides the data features into two different classes: static features and temporal features. Static features $ \textbf{S} $ do not vary over time, e.g., gender, while temporal features $ \textbf{X} $ change. In other words, the static features are characteristics of the data that are not directly related to the time series sample but capture important properties of it.

The \gls{timegan} is tested in four different datasets. First, a synthetic dataset is generated using sinusoidal sequences with an average length of 24 points. A stock prices time series dataset from 2004 to 2019 is used with an average length of 24 days. Third, a dataset of energy prediction with samples of 24 hours of length on average is used. Finally, a medical lung cancer events dataset is used with an average of 58 events per sample.

The proposed architecture adds, in addition to the generator and discriminator networks, two new networks: the encoder and recovery networks. These networks are responsible for embedding the input data in the latent space, as an autoencoder\cite{kramer1991nonlinear} would traditionally do. This system learns the so-called embedding and recovery functions to take the static and temporal features into two separate latent codes $ \textbf{h}_{s} $ and $ \textbf{h}_{t} $ and recover the input information $ \textbf{S}$ and $ \textbf{X} $.

The generator and discriminator parts of the network do the same work as they would do in a traditional \gls{gan}, using the discriminator to differentiate between real and synthetic samples. But in this case, the generator generates the data for the embedding space, while the discriminator also takes this embedding as input for its classification.

The main innovation of \gls{timegan} is implemented in the generator, which, in addition to the normal generation of synthetic samples, is also forced to learn the stepwise dependencies of the data. To do so, the generator receives as input the synthetic embedding $ \textbf{h}_{s}, \textbf{h}_{t-1} $ and computes the next vector $ \textbf{h}_{s}, \textbf{h}_{t} $. This new function is learned by a new supervised loss function that compares the generator forecast with the real data.

\begin{figure}[h]
    \centering
    \includegraphics[width=.4\textwidth]{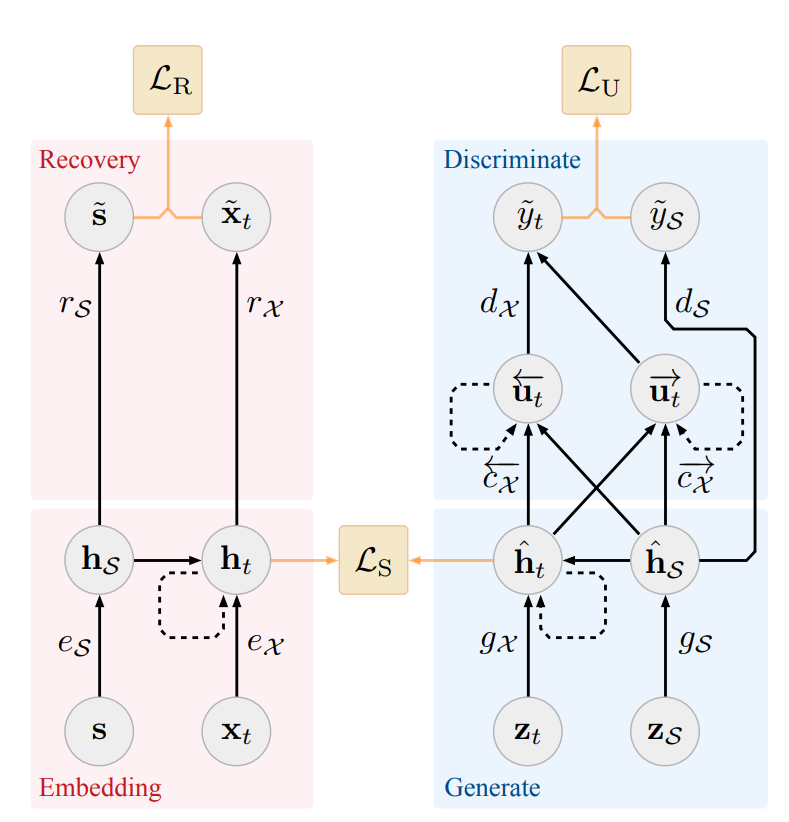}
    \caption{\gls{timegan} architecture. Extracted from \cite{yoon2019time_timegan}.}
    \label{figure:TimeGAN_Architecture}
\end{figure}

Therefore, the training objectives of the presented architecture can be divided into 3 different loss functions.
\begin{itemize}
    \item \textbf{Reconstruction loss} ($ \mathcal{L}_{R} $): This loss is used in the reversible mapping part of the network, composed of the encoder and recovery networks. The length T of each sequence is also a random variable, the distribution of which, for notational convenience, it will be absorbed into distribution p. It is given by:
    \begin{equation}
        \mathcal{L}_{R}=\mathbb{E}_{\textbf{s}, \textbf{x}_{1: T} \sim p}\left[\|\textbf{s}-\tilde{\textbf{s}}\|_{2}+\sum_{t}\left\|\textbf{x}_{t}-\tilde{\textbf{x}}_{t}\right\|_{2}\right],
    \end{equation}
    where the tilde denotes the reconstructed samples and $||\cdot||$ stands for the standard (euclidean) norm.

    \item \textbf{Unsupervised loss} ($ \mathcal{L}_{U} $): In TimeGAN, the generator has two different types of input during training. First, the generator receives synthetic embeddings $\hat{h}_{\mathcal{S}}, \hat{h}_{1:t-1}$, which are autoregressive, to generate the next synthetic vector $\hat{h}_{t}$. In this process, the gradients are computed under unsupervised loss. This is as expected, that is, to allow maximizing D or minimizing G the probability of providing the correct classifications $\hat{y}_{\mathcal{S}}$, $\hat{y}_{1:T}$ for both training data $ \textbf{h}_{s}, \textbf{h}_{1:T} $ and synthetic output $\hat{h}_{\mathcal{S}}, \hat{h}_{1:T}$ from the generator. The unsupervised loss function is the equivalent loss function of a normal \gls{gan} that attempts to distinguish real and fake samples. It is given by:
    \begin{equation}
        \mathcal{L}_{\mathrm{U}}=\mathbb{E}_{\mathrm{s}, \mathbf{x}_{1: T} \sim p}[\log y_{\mathcal{S}}+\sum_{t} \log y_{t}]+\mathbb{E}_{\mathbf{s}, \mathbf{x}_{1: T} \sim \hat{p}}[\log \left(1-\hat{y}_{\mathcal{S}}\right)+\sum_{t} \log \left(1-\hat{y}_{t}\right)]
    \end{equation}
    where $ y_{s} $ and $ y_{t} $ are the classification of the discriminator for static and temporal features and the accent denotes synthetic samples.
    
    \item \textbf{Supervised loss} ($ \mathcal{L}_{S} $): To encourage the generator to learn the conditional transitions of the data, this function is designed that measures the similarity between the real and the synthetic samples created by the generator when applying the forecasting. The loss function is denoted as follows:
    \begin{equation}
        \mathcal{L}_{\mathrm{S}}=\mathbb{E}_{\mathbf{s}, \mathbf{x}_{1: T} \sim p}\left[\sum_{t}\left\|\mathbf{h}_{t}-g_\mathcal{X}\left(\mathbf{h}_{\mathcal{S}}, \mathbf{h}_{t-1}, \mathbf{z}_{t}\right)\right\|_{2}\right]
    \end{equation}
    where $ g_\mathcal{X} $ denotes the sample synthesised by the generator, taking as input the embedded anterior sample $ \mathbf{h}_{\mathcal{S}}, \mathbf{h}_{t-1}, \mathbf{z}_{t} $. An overview of the learning scheme of \gls{timegan} can be seen in figure \ref{figure:TimeGAN_Architecture}.
\end{itemize}

\subsubsection{Conditional Sig-Wasserstein \gls{gan}}
The Conditional Signature Wasserstein \gls{gan} \cite{ni2020conditional} was proposed as a method of maintaining long temporal dependencies in time series data. Regarding the previous models, this architecture is able to outperform previous models such as \gls{timegan} providing better synthetic data for \gls{da}.

In this paper it is proposed a new metric for evaluating the properties of a data stream, providing a description of the sample. This metric is used in the \gls{d} network to differentiate real and synthetic samples. The Sig-Wasserstein metric measures the path space of the data using the Wasserstein distance. In this case the main strength of this method is that it simplifies the training by replacing a neural network \gls{d} for a Linear Regression using the Sig-Wasserstein distance. This process eliminates the cost of approximating a \gls{d}.

The results of the model are tested using different stock market datasets to predict the close prices and the volatility of different actives.

As a \gls{g} network, it is used an AR-FNN generator, which is able of capturing temporal dependencies of time series data.

\begin{figure}[h]
    \centering
    \includegraphics[width=.8\textwidth]{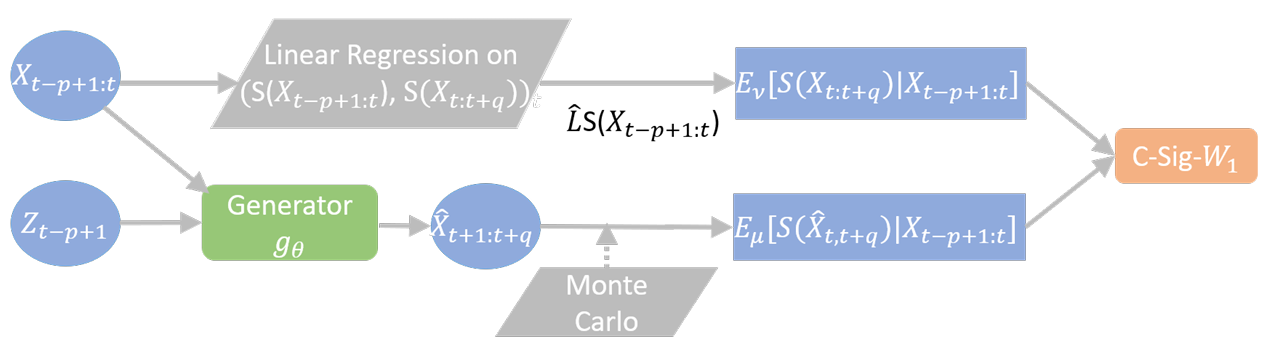}
    \caption{Conditional Sig-Wasserstein \gls{gan} training. Extracted from \cite{ni2020conditional}.}
    \label{figure:CSigWassGAN}
\end{figure}

Figure \ref{figure:CSigWassGAN} shows a scheme of the training process of the Conditional Sig-Wasserstein \gls{gan}. As it can be seen, the real and fake samples can be distinguished by using the Sig-Wassterstein metric.

\glsreset{dtw}
\subsection{\gls{da} based on \gls{dtw}}
\subsubsection{\gls{dtw} Barycenter Averaging}
\gls{dtw}\cite{sakoe1978dynamic} is a classical algorithm that measures the similarity between two data sequences. This method was used as a base in \cite{forestier2017generating}, where the effectiveness of the new generated synthetic time series data was evaluated using augmented training sets for time series classification through the 1-NN classifier in conjunction with \gls{dtw}. In particular 85
datasets of the UCR archive\cite{UCRArchive2018} were used in the experiments. The idea is to manipulate the distribution manifold to generate infinite new samples of data. They achieve this by changing the weights of a set of time series, such as the set $ D=\left\{\left(T_{1}, w_{1}\right), \ldots,\left(T_{N}, w_{n}\right)\right\} $ is embedded in a space $ E $ and the average of \gls{dtw} is denoted as follows:
\begin{equation}
    \arg \min \bar{T} \in E \sum_{i=1}^{N} w_{i} \cdot \textrm{DTW}^{2}\left(\bar{T}, T_{i}\right)
\end{equation}
where $ w $ is the weight of each sample.

To calculate $ \bar{T} $ they use the Expectation-Maximization algorithm and to decide the weight values, three different methods are proposed:
\begin{itemize}
    \item Average All: This method generates the weight vector values using a flat Dirichlet distribution. The main problem with this method is that it tends to fill in data spaces where it should not.
    
    \item Average Selected: This method focusses on selecting a subset of close samples. Thus, it prevents empty spaces from being filled with information because the subsets of samples are close together in the manifold.
    
    \item Average Selected with Distance: The difference between this method and the previous one is that this method calculates the relative distance between the near samples of data.
\end{itemize}

\subsubsection{Suboptimal element alignment averaging}
\gls{spawner}\cite{kamycki2019data} is a \gls{da} method based on the \gls{dtw} algorithm\cite{sakoe1978dynamic}. The \gls{dtw} algorithm is used in this \gls{da} method to align different multidimensional signals $ X_{1}, X_{2} $, giving the so-called \textit{warping path} $ W $ which is a sequence of points that minimises the distance between these input signals. The results of the model were tested using two different \gls{ecg} datasets.

\gls{spawner} algorithm takes the warping path calculated with the \gls{dtw} algorithm and introduces a new random element to the sequence, known as $ w_{p} $. This new point is generated using a uniformly distributed random number within the range $ (0, 1) $. Then the new optimal path is forced to contain the new generated element, obtaining the new warping paths $ W_{1}^{*}, W_{2}^{*}$. Both sequences are aligned using a parameter called $ \xi $, which reduces the flexibility of the path. Finally, both warp paths are concatenated, generating the path $ W_{1,2}^{*} $ from which the new time series signals $ x_{1}^{*}, x_{2}^{*} $ are obtained.

It is observed that for some multivariate signals, this variation of \gls{da} is not enough. Therefore, a random variance is also applied to each point of the signal using a normal distribution such as $x^{*} \sim N(\mu, \sigma^{2}),\mu=0.5(x_{1}^{*}+x_{2}^{*}),\sigma=0.05|x_{1}^{*}-x_{2}^{*}|$.

The use of different alignment methods for text or image data is also proposed, instead of using \gls{dtw} which is proposed when signals are used. Therefore, the overall algorithm can be easily translated to other domains, with the need for an alignment method between two samples.

\section{Discussion}
\label{section:Discussion}
Data augmentation algorithms in the time series domain are really important for improving the available datasets, whose creation is not always easy. In general terms, all the methods presented in this work are algorithms specifically designed for \gls{da} in time series but, in other cases, they are usually adaptations of architectures that were originally designed for other domains, such as image processing. However, the GAN-based algorithms themselves have their beginnings in the field of imaging and have gradually been integrated into other areas.

Regarding the length of the data that each algorithm can process, it should be noted the variety of sizes and dataset types of the different reviewed researches. One of the main strengths that artificial intelligence algorithms have is that, due to the fact that their learning is based on the particular data of each application, they can operate with almost any source of data. This leads to a great variation in the size of the time series windows that most of the algorithms use. In some cases, the data are even processed to treat it as an image, which is symptomatic of the flexibility of \gls{dl} algorithms. In the case of traditional algorithms, they can also process any length of data if it is properly adapted.

In this section, it will be analysed the main advantages and disadvantages of each type of algorithm.

\subsection{Advantages}
Traditional algorithms are widely developed and studied so that their results can be fairly compared. In \gls{da}, they allow you to work by modifying the examples already present, which allows you to control variations. In addition, the simplicity of the algorithms themselves by greatly reducing the number of hyperparameters to be configured results in less time to set them up and the need for less data to train them.

Second, the \gls{vae} generative algorithms allow one to control to a greater extent the variability of the generated data by directly influencing the standard deviation of the latent distribution of the original dataset. This feature allows, among all algorithms, the greatest control of the variability of the generated data. \gls{vae} is commonly used for anomaly detection cases due to its better performance.

Finally, the most current generative models are a breakthrough in the area due to their great results. \gls{gan}s, like \gls{vae}s, allow synthetic data to be generated and, at the cost of losing some control over data generation, they are algorithms capable of much better generalisation. All this is due to the training scheme itself, which allows \gls{gan}s models to learn the distribution that follows the original dataset and, through it, generate synthetic data according to the distribution of the dataset.

Furthermore, since \glspl{gan} are relatively recent algorithms, they benefit from greater attention from the scientific community, which means that there is more recent research focused on improving their results than other algorithms.

\subsection{Disadvantages}
In terms of limitations, the use of traditional algorithms is quite limited because they are based on making modifications to elements of the real dataset. Therefore, they can often produce invalid examples. In general, they are limited to generate examples of lower quality and never to generating new elements. Normally, the reviewed algorithms require a pre-processing phase to normalise the input data, which can lead to more complex algorithms involving previous steps. Otherwise the learning of the \gls{ann} would be inefficient\cite{shao2020normalization, sola1997importance}

Although \gls{vae}s are algorithms capable of generating synthetic data, as opposed to traditional algorithms that only modify the original data, new \gls{nn} models such as \gls{gan}s have mitigated their use in the field because by nature they are capable of generating fewer data than the most current generative networks. Despite this, because they can very precisely control the variability of the generated data, there are fields of application that still continue to use them.

Regarding \glspl{gan}, it can be said that despite their great results, there are certain difficulties that slow down their progress. \gls{gan}s are by far the most complex models currently available and, due to the particularities of the way they are trained, they are extremely difficult to train and obtain results.

\glspl{gan} are one of the most difficult models to train. The main problems that these networks suffer are Mode Collapse \cite{zhang2021mode, adiga2018tradeoff}, instability \cite{goodfellow2020generative}, convergence evaluation \cite{barnett2018convergence} and evaluation metrics.

Due to the \gls{gan} instability problems\cite{arjovsky2017towards, goodfellow2020generative, gonog2019review}, most of the time a great portion of the samples synthesised by the \gls{g} network lack of quality in certain aspects as the emergence of image artefacts \cite{lee2019multi}. In addition, the lack of convergence evaluation of the networks makes it hard to detect when the generated data is of high quality. Therefore, one of the main problems of \glspl{gan} is that their results are not fully reliable.

\subsection{Open issues and challenges}
Some authors \cite{DAvsDG} tend to differentiate between \gls{da} and data generation due to the great advances made in \gls{nn} models. Traditional algorithms are always framed in the area of \gls{da} since the data they produce are always based on existing data; as an open problem, they generate less varied data but more control over what is generated. Furthermore, data generation algorithms produce new data so aggressively that much of the generated data is not possible, degenerating the quality of the augmented dataset \cite{DGdegenerate}.

Unlike the limitation of the scarcity of data augmented with traditional models, \gls{ae}s and \gls{vae}s are born to cover the deficiency of the generated data. In \cite{traditionalvsgenTIMESERIES}, they demonstrate the capability of generative \gls{nn} models to add more diversity to the dataset. In addition, traditional algorithms tend not to be flexible in taking a trained model and applying it to another problem, forcing a rethink of the algorithm. Neural networks, in this aspect, tend to be more flexible, and able to use the same trained model in different problems. In \cite{ramponi2018t}, T-CGAN (Section \ref{TCGAN}) where different datasets are exposed with the same architecture, or in \cite{zhu2019novel}, LSTM-GAN that uses as inputs datasets as disparate as one made of electrocardiograms and another comprising taxi statistics.

However, although generative models offer great advantages, GANs have significant additional problems, especially in training. Typical problems such as modal collapse, Nash equilibria, gradient vanishing or instability are suffered in every training of these models, making their optimisation a very complex process \cite{ganproblems,gonzalez2022improving}.

In general, all generative models share the same open problem that often complicates their validation process. As shown in Section \ref{section:evaluationmetrics}, despite the existence of some evaluation metrics, there is no consensus in the community on which should be used. For example, in \cite{yoon2019time_timegan} authors use empirical evaluation for data generation, but for visualisation they use PCA and a discriminative and a predictive model to see how they have improved after adding the synthetic images. In \cite{donahue2018adversarial_wavegan}, authors propose the Inception Score, a measure of Nearest Neighbour and empirical measurement by humans, and in \cite{huang2021deep},  traditional measures of deep learning (MAE, RMSE and MRE) to compare the generation of future values are used. If the focus is also put on \gls{gan} models, it must be taken into account that, to this problem, there is no method for these architectures to define what the stop condition is in a training.

\section{Conclusion}
\label{section:Conclusion}
Due to the significant evolution that \gls{da} has undergone in recent years, more and more fields are emerging in which to apply and improve the results. This article is focused on giving a comprehensive overview of the main algorithms used for \gls{da} in the field of time series. The review is organised in a taxonomy, consisting of basic and advanced approaches, where it is summarised representative methods of each algorithm (traditional, \gls{vae}s and \gls{gan}s) comparing them empirically, disaggregate by application areas and highlight advantages/disadvantages for future research.

\section*{Acknowledgments}
This work was supported by the Comunidad de Madrid under Convenio Plurianual with the Universidad Politécnica de Madrid in the actuation line of Programa de Excelencia para el Profesorado Universitario. The third-named author has been partially supported by the Madrid Government (Comunidad de Madrid – Spain) under the Multiannual
Agreement with the Universidad Complutense de Madrid in the line Research Incentive for
Young PhDs, in the context of the V PRICIT (Regional Programme of Research and Technological Innovation) through the project PR27/21-029 and by the Ministerio de Ciencia e Innovaci\'on Project PID2021-124440NB-I00 (Spain).

\section{Conflict of Interest}
All authors declare that there is no conflict of interest in this paper.

\section{Data availability}
No datasets were generated or analysed during the current study, and all figures are taken from the papers listed in the references or are generated by us.

\bibliographystyle{unsrt}
\bibliography{sn-bibliography}

\end{document}